%% file: MoHE.tex
\documentclass[sigconf,nonacm]{acmart}

\usepackage{microtype}
\usepackage{graphicx}
\usepackage{hhline}
\usepackage{booktabs} 
\usepackage{array}
\usepackage{makecell}
\usepackage{tabularx}
\usepackage{makecell}
\usepackage{soul}
\usepackage{float}
\usepackage{xcolor} 

\usepackage{amsfonts}
\usepackage{amsthm}
\usepackage{bm}

\usepackage{multirow}
\usepackage{subcaption}
\usepackage{hyperref}
\usepackage{tikz}
\usepackage[most]{tcolorbox}
\usepackage{pifont}
\usepackage{arydshln}
\usepackage[linesnumbered,ruled,vlined]{algorithm2e}
\usepackage{bbold}
\usepackage{wrapfig}
\usepackage{dblfloatfix}
\usepackage{comment}

\newtheorem{definition}{Definition}[subsection]

\begin{document}

\title{Multi-output Headed Ensembles for Product Item Classification}

\author{Hotaka Shiokawa}
\authornote{Corresponding author.}
\email{hotaka.shiokawa@fmr.com}
\affiliation{%
	\institution{Fidelity Investments, Workplace Investing Business Unit}
	\streetaddress{245 Summer Street}
	\city{Boston}
	\state{MA}
	\country{USA}
	\postcode{02210}
}

\author{Pradipto Das}
\authornote{Corresponding author.}
\email{pradipto.das@rakuten.com}
\affiliation{%
	\institution{Rakuten Institute of Technology}
	\streetaddress{2 South Station}
	\city{Boston}
	\state{MA}
	\country{USA}
	\postcode{02110}
}

\author{Arthur R. Toth}
\author{Justin Chiu}
\email{arthur.toth@rakuten.com}
\email{justin.chiu@rakuten.com}
\affiliation{%
		\institution{Rakuten Institute of Technology}
		\streetaddress{2 South Station}
		\city{Boston}
		\state{MA}
		\country{USA}
		\postcode{02110}
}

\renewcommand{\shortauthors}{Shiokawa and Das, et al.}
	
\input{src/abstract}



\maketitle

\input{src/introduction}

\input{src/proposed-models}

\input{src/datasets}

\input{src/experimental-setup}

\input{src/experiments-and-evaluations}

\input{src/conclusion}

\bibliographystyle{ACM-Reference-Format}
\bibliography{MoHE}

\input{src/supplementary-materials}

\end{document}

%% file: src/abstract.tex
\begin{abstract}
In this paper, we revisit the problem of product item classification for large-scale e-commerce catalogs.
The taxonomy of e-commerce catalogs consists of thousands of genres to which are assigned items that are uploaded by merchants on a continuous basis.
The genre assignments by merchants are often wrong but treated as ground truth labels in automatically generated training sets, thus creating a feedback loop that leads to poorer model quality over time.
This problem of taxonomy classification becomes highly pronounced due to the unavailability of sizable curated training sets.

Under such a scenario it is common to combine multiple classifiers to combat poor generalization performance from a single classifier.
We propose an extensible deep learning based classification model framework that benefits from the simplicity and robustness of averaging ensembles and fusion based classifiers.
We are also able to use metadata features and low-level feature engineering to boost classification performance. 
We show these improvements against robust industry standard baseline models that employ hyperparameter optimization.

Additionally, due to continuous insertion, deletion and updates to real-world high-volume e-commerce catalogs, assessing model performance for deployment using A/B testing and/or manual annotation becomes a bottleneck.
To this end, we also propose a novel way to evaluate model performance using user sessions that provides better insights in addition to traditional measures of precision and recall.
    
\end{abstract}

%% file: src/introduction.tex
\section{Introduction}
\label{section:introduction}
	
Product taxonomy classification is a tricky problem to solve for large-scale e-commerce catalogs that are continuously evolving.
The difficulties stem from the following factors: 
\begin{enumerate}
    \item Continuous large-scale manual annotation is infeasible. 
    Data augmentation, semi-supervised and few-shot learning \cite{Ratner-et_al-VLDB-2020, Schick-Schutze-NAACL-2021, Schick-Schutze-EACL-2021, Brown-et_al-NeurIPS-2020} \textit{may} help but with no guarantees.
	    
	\item The efficacy of data augmentation and semi-supervised learning methods gets severely limited in the presence of label noise, which in industrial settings can easily range around 15\% \cite{Shen_et_al-2011}.
	Identifying the nature of corruption in labels is non-trivial \cite{Inouye-2017}.
	Internal assessments reveal that the genre assignment error rate by merchants is around 20\% for the large scale catalog with more than $13K$ leaf nodes in the product taxonomy, which has been made available to us.
	    
	\item There is often an unknown covariate shift \cite{Ioffe-Szegedy-2015,Lipton-et_al-2018} in the final evaluation dataset that consists of the Quality Assurance (QA) team's preferred ways of sampling items including those strategies that provide incentives to merchants.
\end{enumerate}
	
While these problems have no robust solutions, practitioners of taxonomy classification for products in e-commerce catalogs have always relied on improving a \textit{``science''} Key Performance Indicator (science KPI) that is usually based on F1 and AUC scores.
Recently hosted data challenges for this problem \cite{Yiu-Chang_et_al-sigir-dc-ICTIR-2019, Hesam_et_al-sigir-dc-ECIR-2020} and papers in the extreme classification domain \cite{Prabhu-et_al-2018, Prabhu-et_al-2018b, Prabhu-et_al-2020, Saha-et_al-2020, Saini-et_al-2021} highlight the advances in neural network based classification models for product catalogs.

In general, a fusion of classifiers works better than single classifiers. 
However, the \emph{final selection of classifiers for deployment} in production depends on several business Service Level Agreements (SLAs).
Key among them are model performance metrics on rigorous validation datasets that are not affected by covariate shifts, some acceptance criteria by the QA team on the final evaluation dataset and \emph{compute cost} in terms of training and serving time.
The economic cost is usually justified if \emph{significant improvements} in KPIs are obtained.
In \S\ref{section:evaluation}, we highlight a different way of evaluating models that tries to replace the need for expensive manual curation of evaluation sets on a continuous basis and instead rely on privacy preserving user interactions during product search sessions.
	
Effective catalog classifiers have been fastText \cite{Joulin_et_al-fasttext-EACL-2017, Yiu-Chang_et_al-sigir-dc-ICTIR-2019} and more recently fastText with automatic tuning of hyperparameters\footnote{\url{https://fasttext.cc/docs/en/autotune.html}} and Neural Network based ones \cite{Cevahir-Murakami-COLING-2016, Xia-EACL-2017,  Mittal_et_al-WSDM-2021}.
The fastText classifier is chosen for its speed and robustness due to a single linear hidden layer logistic regression model trained with Asynchronous Stochastic Gradient Descent algorithm and feature bucketing.
However, the representational capacity of fastText is limited due to the use of a fixed inventory of possible features and the lack of non-linearities. 
On the other extreme, some large cloud based services provide neural network models that find the best architecture of the network automatically from the data for a particular task.
Popular among these is Google Cloud Platform's AutoML\footnote{\url{https://cloud.google.com/automl}} \cite{Barret_Quoc-NAS-with-RL-ICLR-2017, Liu_et_al-ECCV-2018}.
However, it is difficult to train and deploy such models repeatedly due to the conflicting goals of dollar cost and solution time.
The hypothesis space of estimators beginning with fastText at one end and AutoML at the other consists of hundreds of innovations in designing neural network architectures, self-attention based pre-training and fusion of neural networks \cite{Yiu-Chang_et_al-sigir-dc-ICTIR-2019}.

In \S\ref{section:proposed-models}, we propose a minimalistic Neural Network (NN) architecture \emph{framework} that takes advantage of reduction of estimator variance for ensembles and the advantages of fusing several classifiers.
Ensembles of \emph{independent} estimators generalize better than an individual estimator in that the variance of the ensemble estimator is better than the worst individual estimator.
Let us denote the $T$ independent estimators that estimate the posterior class probabilities by $g^{\mathcal{D}}_t(\mathbf{x})$, where $\mathcal{D}$ is the training dataset and $\mathbf{x}$ is any sample.
Let us also denote the estimator with the \textit{worst variance} to be $g_i(\mathbf{x})$, for some $i \in \{1,...,T\}$, dropping the superscript ${\mathcal{D}}$ where dependence on ${\mathcal{D}}$ is assumed and let this variance be $\sigma^2$.
Then we have,

\vspace{-0.4cm}
\begin{align}
    Var\left(\frac{1}{T}\sum_{t=1}^T g_t(\mathbf{x})\right) 
    = \frac{1}{T^2} \sum_{t=1}^T Var\left(g_t(\mathbf{x})\right) 
    = \frac{1}{T^2} \sum_{t=1}^T \sigma_t^2
    \leq \frac{1}{T} \sigma^2 \label{equ_1}
\end{align}
\vspace{-0.2cm}

While designing our proposed model framework named \texttt{MoHE}, our first inspiration has been the mixture of experts (\texttt{MoE}) \cite{Nowlan_Hinton-NeurIPS-1990,Jacobs_et_al-MoE-NeuralComputation-1991,Eigen_et_al-ICLR-2014,Shazeer_et_al-ICLR-2017,Jiaqi_et_al-Youtube-recommendations-MoE-KDD-2018} model -- see Fig. \ref{figure:MoE}.
The \texttt{MoE} model in the context of a neural network is a system of ``expert'' and gating networks with a selector unit that acts as a multiplexer for stochastically selecting the prediction from the best expert for a given task and input.
However, from a generalization point of view, the \texttt{MoE} classifier has a much looser bound than an ensemble of i.i.d estimators.
If $E_{in}(g)$ denotes the in-sample (training) error and $E_{out}(g)$ denotes the out-of-sample (test) error, then using the union bound of probability, we have the following for \texttt{MoE}:

\vspace{-0.5cm}
\begin{align}
    |E_{in}(g) - E_{out}(g)| > \epsilon \implies & |E_{in}(g_1) - E_{out}(g_1)| > \epsilon \ldots \, \textbf{or} \nonumber \\
    & \ldots \, \textbf{or} \, |E_{in}(g_T) - E_{out}(g_T)| > \epsilon
    \label{equ_2a}
\end{align}

Applying the Hoeffding Inequality \cite{Abu-Mostafa-LFD-2012}, we have

\vspace{-0.4cm}
\begin{align}
    P(|E_{in}(g) - E_{out}(g)|> \epsilon) & \leq \sum_{t=1}^T P(|E_{in}(g)-E_{out}(g)| > \epsilon) \nonumber \\
    & \leq 2T e^{-2\epsilon^2 N}
    \label{equ_2b}
\end{align}
where $N$ is the number of in-sample data points.
Equ. \ref{equ_2b} shows that generalization error bound for \texttt{MoE} can be loose by a factor of $T$.

%% file: src/proposed-models.tex
\section{Multi-output Headed Ensemble (\texttt{MoHE}) Framework}
\label{section:proposed-models}

\begin{wrapfigure}{r}{0.45\columnwidth}
    \vspace{-0.5cm}
    \caption{\small{The \texttt{Aggregator} framework with \texttt{AGG} as a ``\emph{fusion}'' layer. It doesn't share the inputs but share the outputs from the encoders of the estimator threads.}}
    \includegraphics[width=0.45\columnwidth]{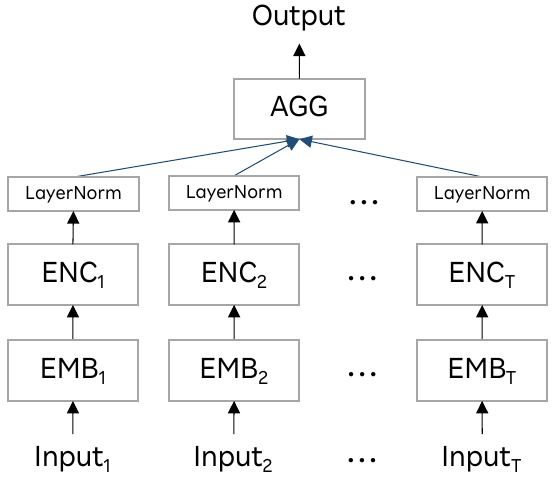}
    \label{figure:aggregator-framework}
    \vspace{-0.5cm}
\end{wrapfigure}

\begin{figure*}
	\centering
	\subcaptionbox{Ensemble\label{figure:ensemble}}{\includegraphics[width=0.17\textwidth]{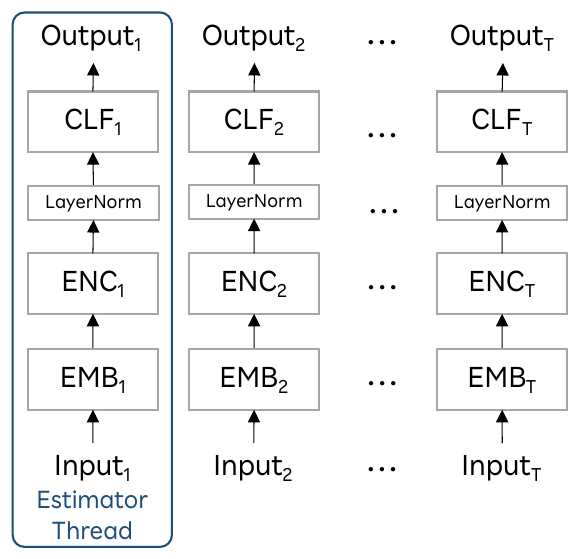}}%
    \hspace{0.01\textwidth}
    \subcaptionbox{MoE\label{figure:MoE}}{\includegraphics[width=0.22\textwidth]{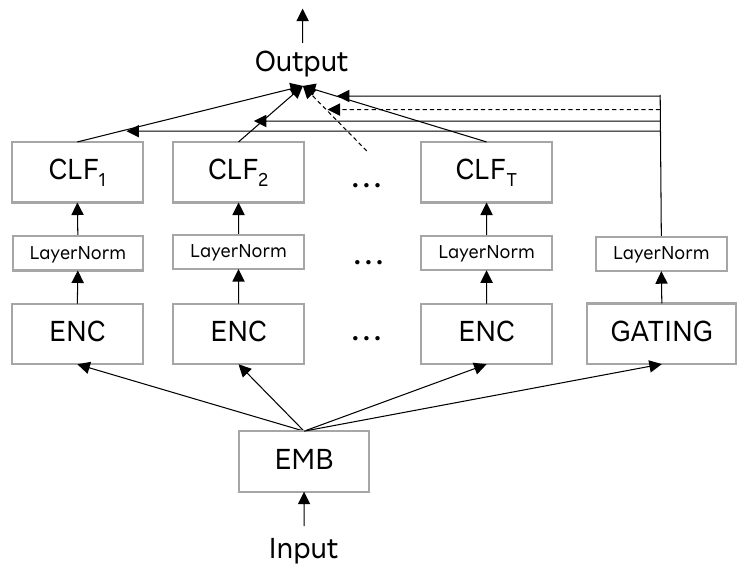}}%
    \hspace{0.01\textwidth}
    \subcaptionbox{MoHE-1\label{figure:MoHE-1}}{\includegraphics[width=0.27\textwidth]{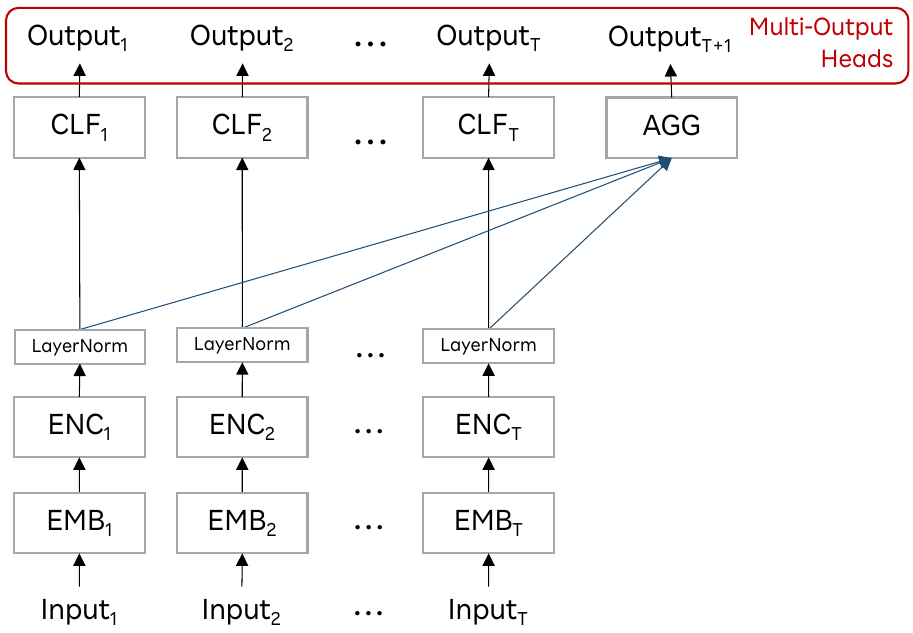}}%
    \hspace{0.01\textwidth}
    \subcaptionbox{MoHE-2\label{figure:MoHE-2}}{\includegraphics[width=0.25\textwidth]{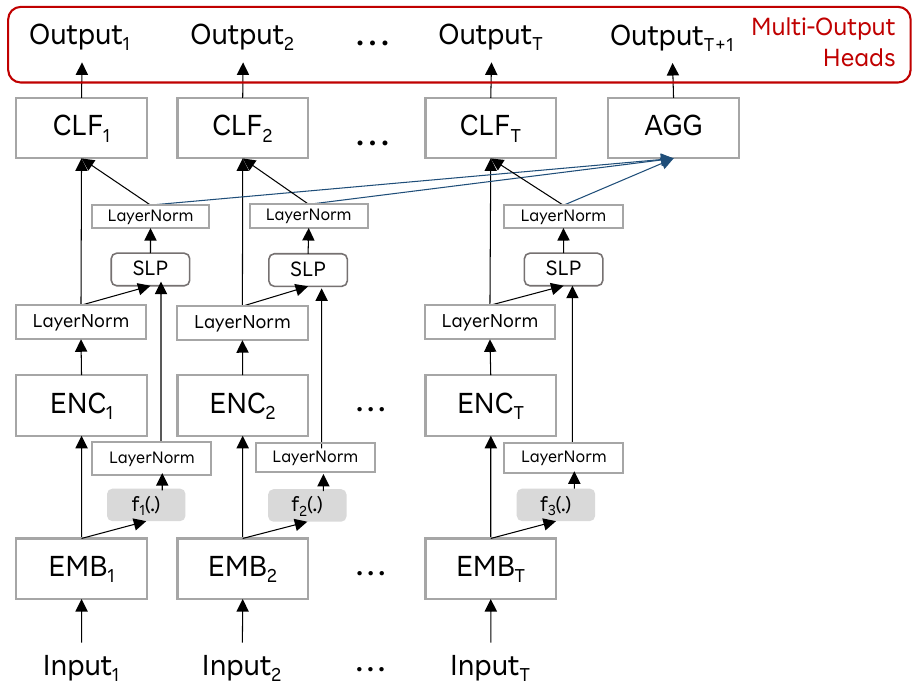}}%
    \caption{Embedding, Encoder and Classifier stacks for different model frameworks. Each stack yields an estimator thread while the set of classifiers that compute posterior class probabilities form the ensemble of \emph{Multi-output} heads.}
	\label{figure:all-models}
\end{figure*}

	
The central idea of \texttt{MoHE} architecture is a loosely coupled ensemble \textit{framework} where each individual classifier's output is considered as a \textbf{\textit{head}}.
This is similar in spirit to the \texttt{BERT} model \cite{Devlin-et_al-NAACL-2019}, however, in this framework, heads are defined only at the output layer.
Furthermore, in this architecture, shown in Fig. \ref{figure:MoHE-1}, any number of independent input-encoder-output units, dubbed \textbf{\textit{estimator threads}} or just \textbf{\textit{threads}} are loosely coupled through an additional classification module dubbed \textbf{\textit{aggregator}}.
The aggregator typically performs the function of a fusion module as shown in Fig. \ref{figure:aggregator-framework}. 

Each thread is allowed to have its own unique (and possibly transformed) input, parameters, encoder and output layer for single task problems.
By design, this framework can easily be extended to handle multi-task problems, but we do not explore that direction.
	
The number of heads, therefore, is $T+1$ where $T$ is the number of threads (estimators) \emph{chosen by design} and the additional one is for the aggregator that loosely couples the estimator threads. 
Posterior class probability estimates can then be obtained by either taking the output from the aggregator alone (see Fig. \ref{figure:aggregator-framework}) or summing all (or part) of the output probabilities from the output heads of the estimator threads including the aggregator. 
The latter usually outperforms the former except at early stages of training or for small training datasets.
	
The analysis of variance for the \texttt{MoHE} framework becomes complicated without any distributional support. 
Let us assume that for each category $k$, the output vector from the heads and the aggregator, $\mathbf{g}_k \equiv \mathbf{g}$, follows multivariate normal distribution.
For a particular head $t$, we can write the covariance and mean for $\mathbf{g}$ as,

\vspace{-0.3cm}
\begin{align}
    \mathbf{g}      =   \begin{bmatrix} 
                        g_{t} \\
                        \mathbf{g}_{\neg t} \\
                        \end{bmatrix}
                    \sim  \mathcal{N} \begin{pmatrix}
                        \begin{bmatrix} 
                        \mu_{g_t} \\
                        \boldsymbol{\mu}_{\mathbf{g}_{\neg t}} \\
                        \end{bmatrix}, 
                        \begin{bmatrix} 
                        \Sigma_{g_t,g_t} & \Sigma_{g_t,\mathbf{g}_{\neg t}} \\
                        \Sigma_{\mathbf{g}_{\neg t},g_t} & \Sigma_{\mathbf{g}_{\neg t}, \mathbf{g}_{\neg t}} \\
                        \end{bmatrix}
                        \end{pmatrix}
    \label{equ_3}
\end{align}
where $\mathbf{g}_{\neg t}$ is a $T$-dimensional vector and $g_t$ is a scalar for each class $k$.
Under this assumption, if we hold all of $\mathbf{g}_{\neg t}$ fixed, we can show that:

\vspace{-0.8cm}
\begin{align}
	\mu_{g_t|\mathbf{g}_{\neg t}} = \mu_{g_t} = \Sigma_{g_t,\mathbf{g}_{\neg t}} \Sigma_{\mathbf{g}_{\neg t},\mathbf{g}_{\neg t}}^{-1} (\mathbf{g}_{\neg t} - \mu_{\mathbf{g}_{\neg t}}) \nonumber \\
	\Sigma_{g_t|\mathbf{g}_{\neg t}} = \Sigma_{g_t,g_t} - \Sigma_{g_t,\mathbf{g}_{\neg t}}\Sigma_{\mathbf{g}_{\neg t},\mathbf{g}_{\neg t}}^{-1}\Sigma_{\mathbf{g}_{\neg t},g_t} = \Sigma_{g_t,g_t} - \hat{\Sigma}_{\mathbf{g}_{T+1}}
	\label{equ_4}
\end{align}

We note that $\Sigma_{\mathbf{g}_{\neg t},\mathbf{g}_{\neg t}}^{-1}$ is positive definite (PD) since $\Sigma_{\mathbf{g}_{\neg t},\mathbf{g}_{\neg t}}$ is.
This is easily shown for an arbitrary PD matrix $\mathbf{A}$ and its eigenvalues $\boldsymbol{\Lambda}$ and eigenvectors $\mathbf{V}$:
$
    \mathbf{A}\mathbf{v} = \lambda_{\mathbf{v}}\mathbf{v} %
    \implies \frac{1}{\lambda_{\mathbf{v}}} \mathbf{v} = \mathbf{A}^{-1} \mathbf{v}
$
for $\lambda_{\mathbf{v}} \in \boldsymbol{\Lambda}$ and $\mathbf{v} \in \mathbf{V}$.
Since $\Sigma_{\mathbf{g}_{\neg t},\mathbf{g}_{\neg t}}^{-1}$ is positive definite and $\Sigma_{g_t,\mathbf{g}_{\neg t}}^{\texttt{T}} = \Sigma_{\mathbf{g}_{\neg t}, g_t}$, hence by definition of positive definiteness that $\mathbf{v}^{\texttt{T}}\mathbf{A}\mathbf{v} \succ 0$,
there is a \textbf{reduction of variance for each} $g_t, t\in\{1,...,T+1\}$ and then Equ. \ref{equ_1} applies.
Note that in Equ. \ref{equ_4}, $\Sigma_{g_t, g_t} \equiv \sigma^2_{g_t}$ for fixed $\mathbf{g}_{\neg t}$.

The caveat for equations \ref{equ_1} and \ref{equ_3} is that they often are oversimplifications in absence of well calibrated class probabilities.
In this paper, we do not tackle calibration, however, it is an important research topic in and of itself -- see \cite{Meelis-NeurIPS-2019} and the references therein.

\subsection{Neural Network Architecture for \texttt{MoHE}}
\label{subsection:formulation}

Our basic \texttt{MoHE} architecture consists of encoder threads with arbitrary parameters and input tokenization. 
The outputs from all encoders (\texttt{CNN}s used here) are globally max-pooled, concatenated, and given to the aggregator module -- see Fig. \ref{figure:MoHE-1} for our \emph{baseline} \texttt{MoHE} architecture. 
We define $T$ to be the number of estimator threads, which are, for instance, independent classifiers in our \emph{baseline} \texttt{ensemble} framework (see Fig. \ref{figure:ensemble}). 
Tokenized input text sequences, $\bm{x}_t$, which can be pre-processed differently for each thread so that $\bm{x}_{t_i} \neq \bm{x}_{t_j}$, are first converted to word embedding vector representations, $\bm{v}_t \in \mathbb{R}^{L_t \times D_t}$, where $L_t$ and $D_t$ are the input text sequence length and embedding dimension, respectively. Denote

\vspace{-0.3cm}
\begin{equation}
    \mathbf{V}_t = f_{t,1}(\mathbf{x}_t) = \texttt{Dropout}\left(\texttt{Embedding}(\mathbf{x}_t)\right)
    \label{equation:V_t}
\end{equation}
where the second index in $f_{t,\cdot}$ refers to the depth in the architecture of the estimator thread.
The subsequent encoding is
\begin{equation}
    \mathbf{u}_t = f_{t,2}(\mathbf{V}_t) = \texttt{Dropout}\left(\texttt{GlobalMaxPool}(\texttt{CNN}_t(\mathbf{V}_t))\right)
    \label{equation:u_t}
\end{equation}
where $\mathbf{u}_t \in \mathbb{R}^{P_t}$ where $P_t$ is the number of filters for $\texttt{CNN}_t$.
We can express the estimator thread, $t$'s output as
\begin{equation}
    \mathbf{g}_t = f_{t,3}(\mathbf{u}_t) = \texttt{Softmax}(\texttt{CLF}_t(\mathbf{u}_t))
    \label{equation:g_t}
\end{equation}
where $\texttt{CLF}_t$ is a densely connected feed forward neural network.
Similarly, the output of the aggregator module is,
\begin{align}\label{equation:g_T_plus_1}
    \mathbf{g}_{T+1} & = f_{T+1, 3}(\{\mathbf{u}_{t \in [1, ..., T]}\}) \\ \nonumber
    & = \texttt{Softmax}\left(\texttt{CLF}_{T+1}\left(\texttt{Concatenate} \left(\mathbf{u}_{t \in [1, ..., T]} \right) \right)\right)
\end{align}

We also apply layer normalization \cite{Ba-et_al-2016} to $\bm{u}_t$ to speed up the convergence and improve performance.
Dropouts \cite{Srivastava-et_al-JMLR-2014} appear as in equations \ref{equation:V_t} and \ref{equation:u_t}.
Contribution to the training loss function for a single data point is

\vspace{-0.4cm}
\begin{equation}
    \mathcal{L} = \gamma_{T+1}\text{CE}(\mathbf{y}, \mathbf{g}_{T+1}) + \sum_{t=1}^{T}{\gamma_t\text{CE}(\mathbf{y}, \mathbf{g}_t)}
    \label{equation:loss}
\end{equation}
where $\mathbf{y}$ is the one-hot representation of a label and $\gamma_{T+1}+\sum_{t=1}^T \gamma_t = 1$ are tuning parameters. 
\emph{The class posterior probabilities to be used for classification could be either} $\mathbf{g}_{T+1}$ or $\frac{1}{T+1}(\mathbf{g}_{T+1} + \sum_{t=1}^T\mathbf{g}_t)$. 
We use the latter and set $\gamma_{T+1}=\gamma_t\, \forall t$ in all of our experiments.
We use Adam optimizer \cite{Kingma-Ba-2014} throughout the paper (except fastText) and do not perform parameter tuning specific to each model to focus on the effects of architectural variations only.

The \texttt{MoHE-2} model (Fig. \ref{figure:MoHE-2}) incorporates additional non-linearities that act as a mini-aggregator module that allows the interaction of information geometries in two spaces -- a function of input's embedding (mean in our minimal framework) and input encoding spaces.   
For this model, Equ. \ref{equation:g_t} becomes:
\begin{equation}
    \mathbf{g}_t = \texttt{Softmax}(\texttt{CLF}_t(\texttt{SLP}(\texttt{Concatenate}(\mathbf{u}_t, \mathbf{V}_t))))
    \label{equation:g_t_MoHE-2}
\end{equation}
where \texttt{SLP} is a single layer perceptron with \emph{tanh} activations.
Equ. \ref{equation:g_T_plus_1} is also changed accordingly for the \texttt{MoHE-2} model.
Dropouts appear after $(\texttt{LayerNorm} \leftarrow f_t(.))$ and $(\texttt{LayerNorm} \leftarrow \texttt{ENC}_t)$ stacks.
	
We use \texttt{CNN} as the encoder in this paper. 
It can be replaced by any other encoders such as RNNs, LSTMs, transformers \cite{Vaswani-et_al-NeurIPS-2017} etc. 
Exploration of arbitrary encoder variations/combinations beyond the scope of this paper.
Furthermore, we restrict ourselves to seven estimator threads and one aggregator module purely for computational reasons of using a single 8-GPU card server.
    
\subsection{Adding \textit{Feature Engineering} to \texttt{MoHE}s}
\label{subsection:MoHE-meta-features}

\begin{wrapfigure}{r}{0.5\columnwidth}
    \vspace{-0.8cm}
    \caption{\small{The threads on the right of estimator thread $T$ are the ``meta estimator threads'' that take as input any desired metadata dubbed \texttt{MetaInput}.
    }}
    \includegraphics[width=0.5\columnwidth]{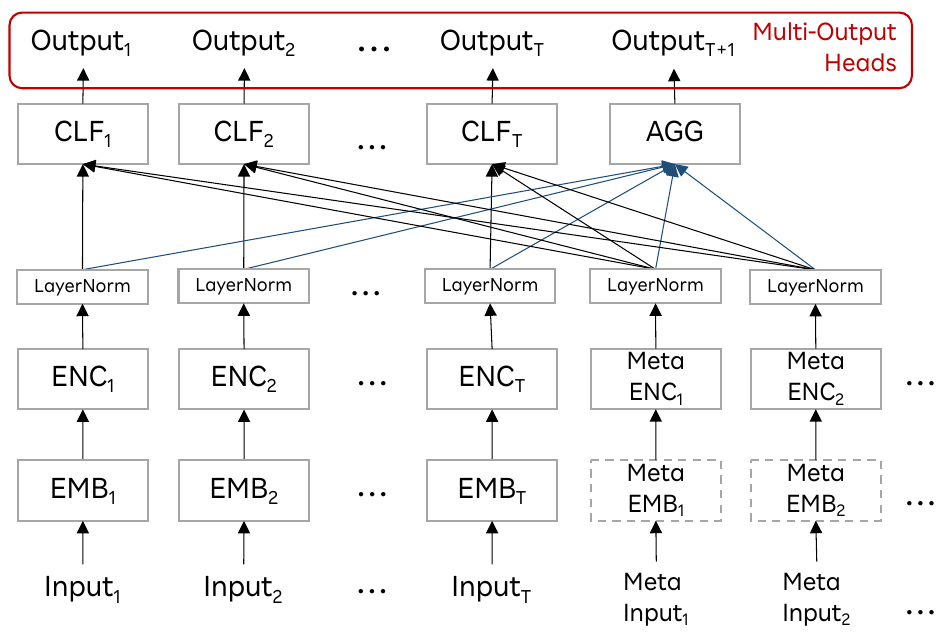}
    \label{figure:MoHE-1-meta-1}
    \vspace{-0.4cm}
\end{wrapfigure}
One of the greatest advantages of the \texttt{MoHE} framework is its ability to accept domain knowledge as additional metadata.
It can add new estimator threads corresponding to individual metadata fields or multiple of them, thus preserving the structure of the data.
On the other hand, if rich meta-data is just appended to main text, forming just another longer text sequence -- as is the case for \texttt{fastText}, it leads to loss of structure and strong coupling of meta-data parameters.

To this end, we feed such auxiliary information, or the products' metadata to the \texttt{MoHE}s in two different ways. 
The first method, dubbed \textit{method-1} hereafter, for \texttt{MoHE-1} is depicted in Fig. \ref{figure:MoHE-1-meta-1}. 
The meta-data inputs are embedded, encoded, and the encodings concatenated with the inputs to all the classifiers (\texttt{CLF} layers) including the aggregator module (\texttt{AGG}). 
Multiple types of metadata could be given to \textit{a single} ``metadata estimator thread'' or to \textit{separate} metadata estimator threads depending on data/encoder types. 
Method-1 is applicable to \texttt{MoHE-2} in exactly the same way as \texttt{MoHE-1}.

In the second method, dubbed \textit{method-2} hereafter, the metadata threads are identical to that of method-1 but their outputs are given to \texttt{SLP}s in the \texttt{MoHE-2} model shown in Fig. \ref{figure:MoHE-2}, instead of directly to the classifiers. 
Method-2 is only applicable for \texttt{MoHE-2}. 
The aggregator module does not take any input from the metadata threads in this case.
We employ basic text (1-dimensional) \texttt{CNN}s with a \emph{kernel size of one} for the metadata encoders. 

\subsubsection{\textbf{Feature Metadata specifics for \texttt{MoHE} Models}}
\label{subsubsection:meta-data-specifics}

The metadata or ``\textit{features}'' that we have used here, appear only in one of the datasets -- a large scale Japanese product catalog.
There are multiple metadata values available for each item, such as various identification numbers related to the products, description, price, ``tags'', image urls, and so on. 
For example, many merchants/shops sell products in only certain categories, and therefore ``\texttt{shop\_ID}'' can be a strong feature for label correlation. 
A similar signal is ``\texttt{tag\_ID}'', that refers to an attribute type of a product.
In this paper, we make use of the maker/brand and shop \texttt{tag}s as features and descriptions as another metadata feature.

As mentioned in \S\ref{subsection:MoHE-meta-features}, the meta estimator threads employ  \texttt{CNN}s with kernel sizes of \emph{one} as their encoders, so as to make them serve as keyword finders. 
For ``descriptions'', however, we keep only nouns, adjectives, and adverbs, and omit repeating words.
The description is thus a sequence of part-of-speech tagged tokens and we set the window size to one here as well. 
This ``\textit{feature engineering}'' of description fits long sentences within a maximum length of $120$.
The length is so set since $90\%$ of the descriptions have length $\leq 120$. 
We use the Mecab\footnote{\url{https://taku910.github.io/mecab/}} tokenizer \cite{Kaji-Kitsuregawa-EMNLP-2014} for tokenizing and extracting parts of speech from Japanese product titles and descriptions.

We show in Table \ref{table:meta-performance} that using metadata for \texttt{MoHE-1/2} models, \emph{performance on the validation set improves by $3\%$ absolute in macro-F1 scores}.
Further, ablation studies show that all three ``\emph{features}'' used together yield best performance.

%% file: src/datasets.tex
\section{Datasets and Preliminary Concepts}
\label{section:datasets}

We first touch upon some definitions that are used throughout this paper.
A \textbf{catalog item} is an item that exists in the product catalog, which is organized as a taxonomy.
All catalog items are assigned to the leaf nodes of the taxonomy tree.

\subsection{Preliminary Concepts}
\label{subsection:definitions}

\begin{definition}
    \textbf{Genre path}: It is the path from the root to a leaf in a product catalog taxonomy. 
    The root is at level $0$ and each node in the path from the root increments the level. 
    Denote each node in the path to be $n_{l}$, where $l$ is a certain $0$-indexed level. 
    The full genre path for an item $\mathbf{x}$ in the product catalog is $n_{0}>...>n_{l}>...>n_{L}$, where $L$ is the leaf level for that path.
\end{definition}
\begin{definition}
    \textbf{Level one genres}: These refer to the genres that are immediately reachable from the root node.
    The names of these genre nodes, (or modifications) usually appear in the front end as ``\emph{departments}'' selection dropdown in a search query box. 
\end{definition}
\begin{definition}
    \textbf{Level one genre path}: Level one or $L1$ genre path is the sequence of nodes identified by $n_{1}>..>n_{l}>..>n_{L}$
\end{definition}
\begin{definition}
    \textbf{Head, Torso, Tail}: Let a genre path be labeled up to level $l$ i.e. $n_{0}>...>n_{l}$.
    Additionally there are no item assignments to any internal node of the taxonomy tree.
    If we sort the $n_l$s in descending order of item counts, then the set $\{n_l\}^{(\mathbf{Head})}$ are those $n_l$ that cover $70\%$ of all items that belong to all nodes at level $l$.
    The set $\{n_l\}^{(\mathbf{Torso})}$ are those nodes, $n_l$, that cover the next $20\%$ of items and the set $\{n_l\}^{(\mathbf{Tail})}$ are those nodes, $n_l$, that cover the last $10\%$ of items.
    For our experiments we set $l=1$ to obtain head, tail and torso segments corresponding to level one genres in the training set.
    See \S\ref{section:supplementary-materials} for a plot of the histograms.
\end{definition}
\begin{definition}
    \textbf{User session}: 
    For a given time period, the product catalog is a static catalog of items with possible updates to quantities and prices.
    This catalog becomes highly dynamic when the user interface of the search engine for the product items record user interactions for \emph{each} item that is retrieved through search queries $Q$.
    These interactions result in time slices of activities per user called \emph{User sessions}.
\end{definition}

\subsection{Datasets}
\label{subsection:datasets}

For our experiments, we use two datasets -- a Japanese product catalog with item metadata from a major e-commerce company in Japan, \texttt{Rakuten Ichiba}, and an English product catalog that was released as part of a data challenge in SIGIR 2018 \cite{Yiu-Chang_et_al-sigir-dc-ICTIR-2019}.
The latter dataset does not have any metadata.
We dub the first dataset the \emph{Rakuten Ichiba} dataset, which is a sample from the full catalog.
This dataset is almost entirely in Japanese. English words and characters, however, do appear sometimes.
There are $38$ level one genres with $\approx 13K$ leaf nodes.
Some statistics of the training data size are shown in Table \ref{tab:baseline_performance_1level}.
The training, validation and evaluation datasets have $23.7M$, $2.96M$ and $1.18M$ items respectively.
Please refer to \S\ref{section:supplementary-materials} for frequency distribution of items in the level one genres.

\begin{wrapfigure}{r}{0.4\columnwidth}
    \vspace{-0.8cm}
    \caption{\small{Dynamic catalog. Attaching User sessions to each catalog item.}}
    \includegraphics[width=0.4\columnwidth]{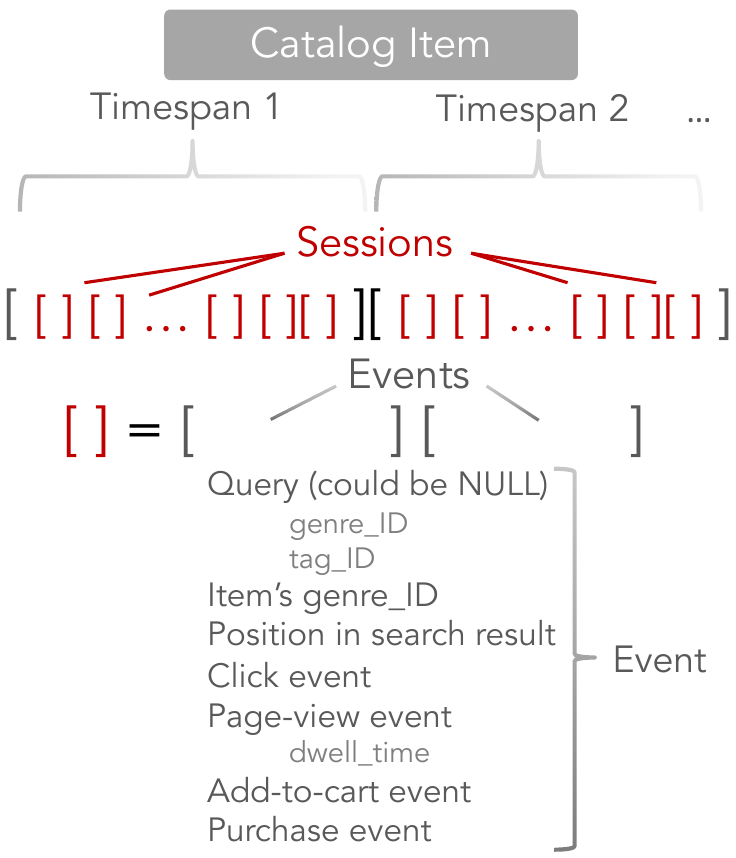}
    \label{figure:user-session}
    \vspace{-0.8cm}
\end{wrapfigure}
Since the Rakuten Ichiba dataset is much larger in size, training a flat classifier is often suboptimal.
We thus deploy a two level classifier similar to that used in \cite{Das-Et_al-EACL-2017}.
We only report scores from the $38$ level one classifiers.

Attached to each catalog item is a list of privacy preserved sessions, each of which belong to a particular user.
Each session is defined over a time span and records events such as the queries issued, search results for the queries, clicked items in the results, page views, whether the clicked item was added to cart and purchased, etc. -- see Fig. \ref{figure:user-session}.
Not all fields, \textit{e.g.} \texttt{dwell\_time}, are made available to us due to GDPR\footnote{\url{https://gdpr.eu/}} restrictions.
Query attribute fields such as \texttt{tag\_ID} are mostly missing and hence not used in experiments in \S\ref{subsection:evaluations-user-interactions}.
	
\texttt{Rakuten} hosted a product taxonomy classification data challenge as part of the SIGIR'18 E-Com Workshop.
The details of the data challenge and the dataset used are mentioned in \cite{Yiu-Chang_et_al-sigir-dc-ICTIR-2019}, and we do not repeat it here.
We utilize that dataset, which is in English, as our second dataset to compare models.
The SIGIR'18 E-Com Data Challenge dataset, dubbed \emph{SIGIR'18-DC} dataset henceforth, is a relatively small dataset of catalog items from \url{www.rakuten.com} consisting of $800K$ training and $200K$ evaluation items.
The labels of the items are arranged in an anonymized taxonomy tree consisting of a total of $3008$ leaf nodes.
For this dataset, we use a flat classification scheme similar to what has been used in \cite{Yiu-Chang_et_al-sigir-dc-ICTIR-2019}.

%% file: src/experimental-setup.tex
\begin{figure*}[!b]
	\centering
	\subcaptionbox{Macro-F1 values plotted against the number of threads for the head genres.\label{figure:nthread_vs_macro_head}}{\includegraphics[width=0.24\textwidth]{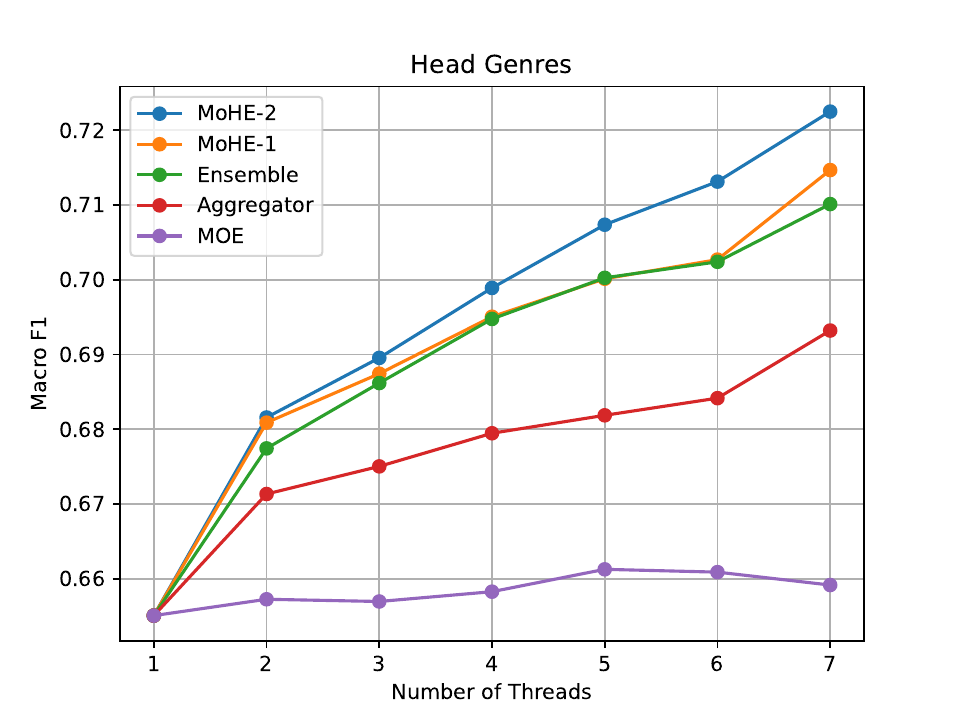}}%
    \hspace{0.05\textwidth}
    \subcaptionbox{Macro-F1 values plotted against the number of threads for the torso genres.\label{figure:nthread_vs_macro_torso}}{\includegraphics[width=0.24\textwidth]{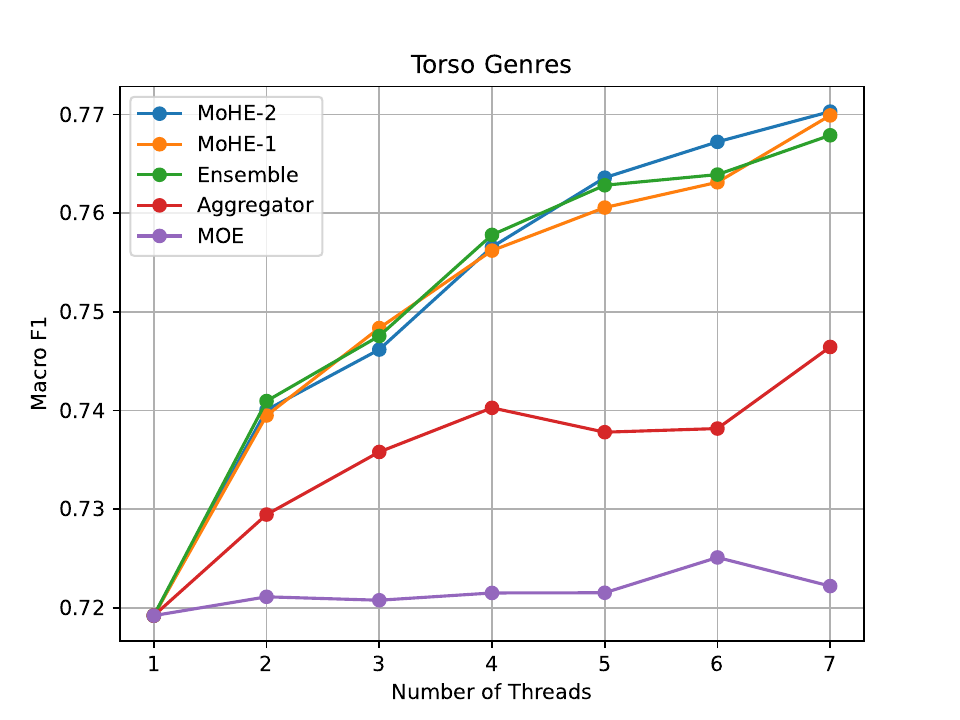}}%
    \hspace{0.05\textwidth}
    \subcaptionbox{Macro-F1 values plotted against the number of threads for the tail genres.\label{figure:nthread_vs_macro_tail}}{\includegraphics[width=0.24\textwidth]{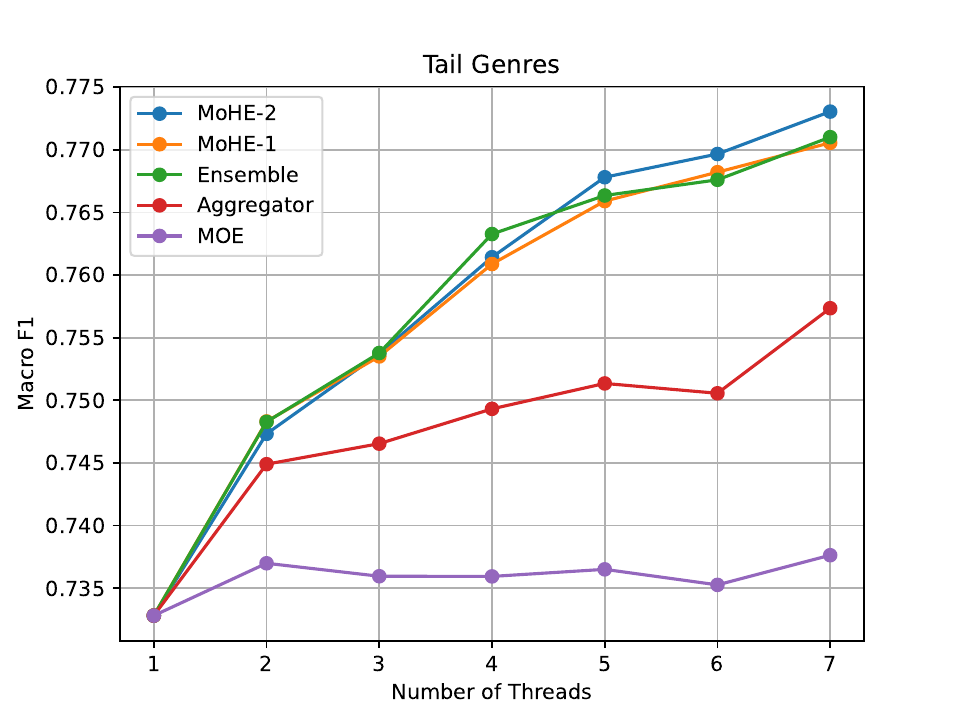}}
    \caption{\small{Plots of Macro-F1 values for the \texttt{MoE}, \texttt{Aggregator}, \texttt{Ensemble}, \texttt{MoHE-1} and \texttt{MoHE-2} models from level one genre path classifiers for the Rakuten Ichiba dataset. The leaf nodes for classification correspond to the level one genres, which are organized into head, torso and tail segments. Overall, there are $38$ level one genres and hence $38$ \emph{groups} of level one classifiers. Each of the $38$ groups represents a set of estimator threads corresponding to a particular classification framework.}}
    \label{figure:head-tail-torso-comparison-for-different-threads}
\end{figure*}

\section{Experimental Setup}
\label{section:experimental-setup}

For all experiments, the Rakuten Ichiba dataset has been partitioned into training, development and \textit{validation} sets, all of which are sampled from the same data distribution.
This distribution of items \textbf{has no sampling bias} in terms of purchase behavior and includes a large sample of items from purchased and non-purchased items and a minor percentage of historical curated items whose genres have been manually corrected.
The data has noisy labels to the extent of $20\%$ based on internal assessment.
We use a \emph{sampling of genres based on purchased items from user sessions to validate this figure of $20\%$} in \S\ref{subsection:evaluations-user-interactions}.

We also have a non-overlapping \textit{evaluation} set for the Rakuten Ichiba dataset, where annotators have sampled items based on GMS (Gross Merchandise Sale) values and corrected the mis-predicted genres from a previous model.
However, for all experiments with the Rakuten Ichiba dataset in this section, we only use the validation set for model comparison.
For the SIGIR'18-DC dataset, the challenge evaluation set is the set of $200K$ items that is mentioned in \cite{Yiu-Chang_et_al-sigir-dc-ICTIR-2019}. 

\begin{table}[!t]
    \small
	\centering
	\caption{\small{Baseline thread parameters. The thread indices are ordered from left to right as shown in Fig. \ref{figure:all-models}. The input sequence lengths are set to $60$ for word based tokenization and $100$ for character based tokenization since greater than $90\%$ of titles are shorter than $60$ words and $100$ characters in length. For the SIGIR'18-DC dataset, the default settings of \texttt{CNN} Kernel sizes for character tokenization is smaller since the average length of English words is $\approx 5$ characters and we use sequential multiples of $5$. ``bi-grams'' is by tokens. }}
	\vspace{-0.3cm}
	\begin{tabular*}{\textwidth}{p{2cm}|p{0.45cm}|p{0.45cm}|p{0.45cm}|p{0.4cm}|p{0.4cm}|p{0.4cm}|p{1cm}}
	\cline{1-8}
	Threads & 1 & 2 & 3 & 4 & 5 & 6 & 7 \\
	\cline{1-8}
	Tokenization & word & word & word & char & char & char & bi-gram \\ \cdashline{1-8}[.4pt/1pt]
	\texttt{CNN} Kernel Size \scriptsize{[Rakuten Ichiba]} & 3 & 4 & 5 & 5 & 15 & 25 & 3 \\ \cdashline{2-8}[.5pt/1.5pt]
	\texttt{CNN} Kernel Size \scriptsize{[SIGIR'18 - DC]} & 3 & 4 & 5 & 5 & 10 & 15 & 3 \\ \cdashline{1-8}[.4pt/1pt]
	Number of Filters & 100 & 100 & 100 & 300 & 300 & 300 & 100 \\ \cdashline{1-8}[.4pt/1pt]
	Input Length & 60 & 60 & 60 & 100 & 100 & 100 & 60 \\
	\cline{1-8}
	\end{tabular*}
	\label{table:baseline-MoHE-parameters}
	\vspace{-0.3cm}
\end{table}

\begin{table}[ht!]
    \scriptsize
    \centering
    \caption{\small{\texttt{BERT} vs. \texttt{MoHE-2} comparison on $4$ randomly selected $L1$ genres from Rakuten Ichiba dataset and full SIGIR'18 E-Com dataset. Bold numbers denote cases where \texttt{MoHE-2} significantly outperforms \texttt{BERT} at $95\%$ confidence interval for bootstrap sampling \cite{Yiu-Chang_et_al-sigir-dc-ICTIR-2019}.}}
    \begin{tabular}{|c|c|c|c|}
    \hline
    \makecell{\textbf{Categories}} & \makecell{\textbf{Comparison parameters}} & \makecell{\texttt{\textbf{BERT}}} & \makecell{\texttt{\textbf{MoHE-2}}}\\
    \hline
    \multicolumn{4}{c}{\emph{Rakuten Ichiba Product Catalog Dataset}} \\
    \hline
	\makecell[l]{\textbf{Musical Instruments}\\\scriptsize{\# items: 162472}\\\scriptsize{\# classes: 295}} & \makecell[l]{Macro-F1\\Training Time (hrs)\\Number of Parameters (M)} & \makecell{0.743\\7.0\\110.2}& \makecell{0.743\\\textbf{0.1}\\\textbf{22.9}}\\
	\cdashline{1-4}[.4pt/.1pt]
	\makecell[l]{\textbf{Beverages}\\\scriptsize{\# items: 187629}\\\scriptsize{\# classes 42}} & \makecell[l]{Macro-F1\\Training Time (hrs)\\Number of Parameters (M)} & \makecell{0.808\\7.3\\110.0}& \makecell{\textbf{0.816}\\\textbf{0.1}\\\textbf{4.9}}\\
	\cdashline{1-4}[.4pt/1pt]
	\makecell[l]{\textbf{Shoes}\\\scriptsize{\# items: 410691}\\\scriptsize{\# classes: 83}} & \makecell[l]{Macro-F1\\Training Time (hrs)\\Number of Parameters (M)} & \makecell{0.712\\15.3\\110.1}& \makecell{\textbf{0.733}\\\textbf{0.3}\\\textbf{17.5}}\\
	\cdashline{1-4}[.4pt/1pt]
	\makecell[l]{\textbf{Computers \& Networking}\\\scriptsize{\# items: 462906}\\\scriptsize{\# classes: 270}} & \makecell[l]{Macro-F1\\Training Time (hrs)\\Number of Parameters (M)} & \makecell{0.694\\20.0\\110.2}& \makecell{\textbf{0.707}\\\textbf{0.3}\\\textbf{47.0}}\\
	\hline
	\multicolumn{4}{c}{\emph{Rakuten.com SIGIR’18 E-Com Workshop Data Challenge Dataset}} \\
	\hline
	\makecell[l]{\textbf{All categories}\\\scriptsize{\# items: 712761}\\\scriptsize{\# classes: 3008}} & \makecell[l]{Macro-F1\\Training Time (hrs)\\Number of Parameters (M)} & \makecell{0.433\\32.0\\112.3}& \makecell{\textbf{0.472}\\\textbf{1.8}\\\textbf{65.1}}\\
	\hline
    \end{tabular}
    \vspace{-0.5cm}
    \label{table:BERT-vs-MoHE-2}
\end{table}

\begin{table*}[!t]
    \scriptsize
	\centering
	\caption{\small{Baseline model performance comparison (Micro-F1/Macro-F1) for the representative nine genres from the \textit{validation set}. \texttt{GCP AutoML}$^*$ ignores rare categories while training. The support set for categories during its evaluation is thus smaller leading to higher Micro-F1 scores being reported by \texttt{GCP AutoML}. The numbers in bold for the \texttt{MoHE-2} column are statistically significant \emph{to both} \texttt{fastText Autotune NNI} and \texttt{Ensemble} under Bootstrap Sampling test with $95\%$ confidence interval as used in \cite{Yiu-Chang_et_al-sigir-dc-ICTIR-2019}. Note that \texttt{MoE}, \texttt{Aggregator}, \texttt{fastText}, \texttt{Ensemble}, \texttt{MoHE-1} and \texttt{MoHE-2} are \emph{not tuned} to individual genres.}}
	\vspace{-0.3cm}

\input{src/tables/table-sample-Ichiba-and-SIGIR}
	\label{tab:baseline_performance_1level}
	\vspace{-0.2cm}
\end{table*}


\paragraph{\textbf{Configurations for \texttt{MoHE} Threads}}

Each estimator thread $t$ is an embedding, encoder and classifier stack with output layer $\mathbf{g}_t$: 
\vspace{-0.1cm}
\begin{align}
    \mathbf{g}_t = \texttt{CLF}_{t,3} ( \texttt{LayerNorm} ( \texttt{ENC}_{t,2} ( \texttt{EMB}_{t,1} (\mathbf{x}) ) ) )
\end{align}
Each thread has different parameters and input tokenization types as summarized in Table \ref{table:baseline-MoHE-parameters}. 
The parameter values are obtained using minimal manual tuning over a development set for our \texttt{Ensemble} model.
The word embedding dimension is set to $\min(\frac{C}{2},100)$ where $C$ is the number of leaf nodes for each level one genre. 
This setting substantially reduces the number of parameters and is set following baseline models mentioned in \cite{Yiu-Chang_et_al-sigir-dc-ICTIR-2019}.
This embedding dimension is set for every model framework except \texttt{fastText Autotune NNI} and \texttt{GCP AutoML}. 
Finally, the dropout values are set to $0.1$.
We experiment with incrementally adding seven estimator threads to all model architectures.
The results are shown in Fig. \ref{figure:head-tail-torso-comparison-for-different-threads}.

The baseline configurations are used for building models for both Rakuten Ichiba and SIGIR'18-DC datasets.
In this paper, we \emph{do not tune} the parameters/properties of the estimator threads for each genre.
Tuning can be performed using many publicly available software packages for neural networks\footnote{Microsoft's Neural Network Intelligence (NNI): \url{https://github.com/Microsoft/nni} and Ray Tune: \url{https://docs.ray.io/en/latest/tune/index.html}}.

%% file: src/tables/table-sample-Ichiba-and-SIGIR.tex
\begin{tabular*}{\textwidth}{p{3.2cm}|p{1cm}|p{1cm}||p{1cm} p{1cm} p{1.5cm} p{1cm} p{1.4cm} p{1cm} p{0.8cm} p{0.8cm}}
\cline{1-11}
Categories  &   Training data size  &   Number of leaf nodes    &   \texttt{MoE} &   \texttt{Aggregator}  &   \texttt{GCP AutoML}$^*$  &   \texttt{fastText}    &   \texttt{fastText Autotune NNI}   &   \texttt{Ensemble}    &   \texttt{MoHE-1}  &   \texttt{MoHE-2} \\ \hhline{|-|--||--------|}
\multicolumn{11}{|c|}{\textbf{Sampling of categories from Rakuten Ichiba Product Catalog Dataset}} \\
\multicolumn{3}{|c}{} & \multicolumn{8}{c|}{\textit{Head (Top three of eleven level one genres ordered by item counts)}} \\
Books   &   \makecell[r]{2927065}  &   \makecell[r]{833}                                    & 0.694/0.593 & 0.720/0.628 & 0.374$^*$/NA & 0.735/0.619 & 0.737/0.611 & 0.739/0.649 & 0.740/0.655 & 0.755/\textbf{0.674}  \\ \cdashline{1-11}[.5pt/1.5pt]
Household Goods, Stationery \& Craft   &   \makecell[r]{2311231}   &   \makecell[r]{596}    & 0.820/0.753 & 0.850/0.791 & 0.670$^*$/NA & 0.857/0.794 & 0.857/0.797 & 0.855/0.797 & 0.856/0.798 & 0.862/\textbf{0.808}  \\ \cdashline{1-11}[.5pt/1.5pt]
Flowers, Gardening and DIY  &   \makecell[r]{1989608}   &   \makecell[r]{776}              & 0.738/0.579 & 0.791/0.630 & 0.575$^*$/NA & 0.797/0.608 & 0.800/0.618 & 0.794/0.639 & 0.794/0.637 & 0.801/\textbf{0.653}   \\ \cline{1-11}
\multicolumn{3}{|c}{} & \multicolumn{8}{c|}{\textit{Torso (Top three of nine level one genres ordered by item counts)}} \\ 
Smartphones and Tablet PCs  &   \makecell[r]{726504 }   &   \makecell[r]{45}                & 0.969/0.784 & 0.973/0.810 & 0.970$^*$/NA & 0.974/0.815 & 0.974/0.822 & 0.975/0.829 & 0.976/0.835 & 0.976/\textbf{0.836} \\ \cdashline{1-11}[.5pt/1.5pt]
Women's Fashion &   \makecell[r]{572892 }  &   \makecell[r]{105}                            & 0.899/0.791 & 0.913/0.815 & 0.903$^*$/NA & 0.912/0.821 & 0.915/0.832 & 0.922/0.824 & 0.922/0.833 & 0.923/0.831 \\ \cdashline{1-11}[.5pt/1.5pt]
Kids, Baby and Maternity    &   \makecell[r]{541637 }   &   \makecell[r]{464}              & 0.851/0.734 & 0.869/0.766 & 0.846$^*$/NA & 0.872/0.753 & 0.868/0.764 & 0.882/0.778 & 0.881/0.786 & 0.882/\textbf{0.782}   \\ \cline{1-11}
\multicolumn{3}{|c}{} & \multicolumn{8}{c|}{\textit{Tail (Top three of eighteen level one genres ordered by item counts)}} \\ 
Shoes  & \makecell[r]{340078 }  &   \makecell[r]{80}                                        & 0.789/0.681 & 0.814/0.707 & 0.797$^*$/NA & 0.824/0.722 & 0.829/0.731 & 0.829/{\underline{0.725}} & 0.829/0.728 & 0.829/{\textbf{0.733}} \\ \cdashline{1-11}[.5pt/1.5pt]
Appliances  &   \makecell[r]{324940 }   &   \makecell[r]{323}                               & 0.831/0.742 & 0.852/0.771 & 0.843$^*$/NA & 0.854/0.764 & 0.860/{\underline{0.772}} & 0.864/0.781 & 0.862/0.778 & 0.863/{\textbf{0.778}} \\ \cdashline{1-11}[.5pt/1.5pt]
Diet and Health &   \makecell[r]{273066 }   &   \makecell[r]{361}                          & 0.761/0.689 & 0.777/0.706 & 0.747$^*$/NA & 0.780/0.673 & 0.782/{\underline{0.704}} & 0.797/0.732 & 0.796/0.725 & 0.797/{\textbf{0.730}} \\ \cline{1-11}
\multicolumn{11}{c}{\textbf{Rakuten.com SIGIR'18 E-Com Workshop Data Challenge Dataset}} \\ 
All categories & \makecell[r]{712761} & \makecell[r]{3008}                                  & 0.780/0.392 & 0.805/0.443 & 0.841$^*$/NA & 0.795/0.395 & 0.804/0.446 & 0.816/0.458 & 0.818/0.461 & 0.820/\textbf{0.472} \\ \cline{1-11}
\end{tabular*}

%% file: src/experiments-and-evaluations.tex
\section{Experiments and Evaluation}
\label{section:evaluation}

Macro-F1 scores induce equal weighting of genre performance and hence are a much stricter standard and that is the measure that we use henceforth for comparing models.
For all models except \texttt{AutoML} and \texttt{fastText NNI}, the scores reported are averages of five runs.
We note that our \texttt{Ensemble} of \texttt{CNN}s baseline (i.e. \texttt{MoHE} without the coupling) is a strong classifier and outperforms \texttt{MoE} and \texttt{Aggregator} baselines significantly.
Additionally, \texttt{MoHE-2} outperforms \texttt{Ensemble} significantly on the validation set.

\subsection{\texttt{BERT} vs. \texttt{MoHE-2}}
\label{subsection:BERT-vs-MoHE-2}

Transformer based deep learning models involving self-attention and large-scale pre-training are used almost ubiquitously now-a-days to solve various language tasks. 
To this end, we compare the \texttt{MoHE-2} framework with \texttt{BERT} \cite{Devlin-et_al-NAACL-2019} for a preliminary comparison on randomly selected $10\%$ of level one genres from the Rakuten Ichiba dataset and all genres from the SIGIR'18 E-Com dataset.
Table \ref{table:BERT-vs-MoHE-2} shows that \texttt{MoHE-2} \emph{significantly outperformed} \texttt{BERT} for most genres along all aspects of Macro-F1, compute time and model size.
Details of the \texttt{BERT} model hyperparameters are mentioned in \S\ref{section:supplementary-materials}.

The main issue with \texttt{BERT} is that it is a more generalized multi-task model where fine-tuning is dependent on a large-scale language model, which is trained according to specific objectives of next word prediction based on a suitably chosen context.
For the case of classification of item titles, the NSP (Next Sentence Prediction) objective of BERT is irrelevant if we are to even pre-train on item titles and so is SOP (Sentence Order Prediction) of \texttt{AlBERT} \cite{Lan-et_al-ICLR-2020}.
\texttt{RoBERTa} \cite{Liu-et_al-ArXiv-2019} removes the NSP objective, however, its training time does not meet our business SLA.
Hence, we drop \texttt{BERT} and similar models from future comparisons in this paper.


\subsection{Effect of Number of Estimator Threads}
\label{subsection:performance-for-number-of-threads}

We start with analyzing the importance of adding successive estimator threads to the model frameworks and compare the graphs in the three plots shown in Figs. \ref{figure:nthread_vs_macro_head}, \ref{figure:nthread_vs_macro_torso} and \ref{figure:nthread_vs_macro_tail}.
As mentioned in \S\ref{section:introduction}, the \texttt{MoE} model is still \emph{a single classifier and has a loose generalization bound} and it performs worst amongst all models compared.

Based on the original \texttt{MoE} model in \cite{Jacobs_et_al-MoE-NeuralComputation-1991}, we can only use one type of input that is shared with the ``\textit{experts}'' and the ``\textit{gate}'' and we choose the configuration shown for estimator thread $1$ in Table \ref{table:baseline-MoHE-parameters}.
Because of this constraint, it also doesn't show much variation in performance since the estimator threads differ only in random initialization of the input embedding.
\texttt{MoE} thus \emph{suffers from bias in input selection} that may also explain its poor performance.
The classification performance shown in Fig. \ref{figure:head-tail-torso-comparison-for-different-threads} with regards to Macro-F1 scores for the level one genre paths of the head segment is overwhelmingly dominated by the \texttt{MoHE-2} model.
For the level one genre paths that belong to the torso and tail segments, \texttt{MoHE-2} also outperforms \texttt{Ensemble} at seven estimator threads.
The additional mini-aggregators introduced in \texttt{MoHE-2} show improvements.

Classification using only the \texttt{Aggregator} module of the \texttt{MoHE} models is an improvement over the \texttt{MoE} model where all the ``expert'' decisions are \emph{fused}.
As mentioned in \S\ref{subsection:formulation}, we set the number of estimator threads to $7$.

\subsection{Model Comparison for Selected Genres}
\label{subsection:model-comparison-for-selected-genres}

Comparison of \texttt{MoHE} with \texttt{MoE} \cite{Jacobs_et_al-MoE-NeuralComputation-1991}, \texttt{Aggregator} Framework (Fig. \ref{figure:aggregator-framework}), \texttt{GCP AutoML}, \texttt{fastText} \cite{Joulin_et_al-fasttext-EACL-2017}, \texttt{fastText Autotuned with NNI} and finally the \texttt{Ensemble} framework is shown in Table \ref{tab:baseline_performance_1level}.
We first sort the level one genres in descending order of item frequency and segment them into head, torso and tail segments. 
We then choose nine categories -- the largest three, each from head, torso and tail segments.
We compare against industry standard \texttt{GCP AutoML} and \texttt{fastText} tuned with Microsoft's NNI.
The nine categories have been chosen to run \texttt{GCP AutoML} within the budget allotted to us.
We run \texttt{GCP AutoML} for at most a day for each of the nine genres.
As of this writing, \texttt{fastText Autotune}\footnote{\url{https://fasttext.cc/docs/en/autotune.html}} is not stable for the larger Rakuten Ichiba dataset.
Out of the box, \texttt{GCP AutoML} constrains the volume of data ingestion, including skipping rare categories thereby hindering apples-to-apples comparison.
It also reports Micro-F1 scores in batch mode and obtaining Macro-F1 scores incur additional cost and thus we don't report them in Table \ref{tab:baseline_performance_1level}.
We drop \texttt{GCP AutoML} from further comparisons.

We find that our proposed \texttt{MoHE} frameworks with our default setting of parameters (see Table \ref{table:baseline-MoHE-parameters}) often perform better than other baselines despite the fact that they consist of lightweight \texttt{CNN} architectures \emph{without being tuned} for a specific genre or dataset.
The gains are obtained more for the head and torso genres and since we do not specifically model category imbalance, the performance on the tail categories are \textbf{not} significantly better \textbf{to both} \texttt{fastText Autotune NNI} and \texttt{Ensemble} but to \underline{the underlined one}.
The performance of \texttt{MoHE-2} model is even better for the SIGIR'18 E-Com dataset that has much less label noise and lower number of classes.

\subsection{Summary of Evaluations for all Genres}
\label{subsect:evaluations-standard-summary}

In this section, we briefly \textit{summarize} the quantitative evaluations for the models and frameworks mentioned in this paper -- \texttt{MoE} model, \texttt{Aggregator} framework, \texttt{fastText Autotune NNI} model, \texttt{Ensemble} framework, \texttt{MoHE-1} framework and \texttt{MoHE-2} framework.
We compare the \texttt{MoHE} frameworks without adding metadata for the Rakuten Ichiba dataset to be fair to the SIGIR'18 E-Com dataset, which does not carry any metadata.

\begin{table}[h!]
    \small
    \centering
    \caption{\small{Macro-F1 scores from the classifiers discussed here on the \textit{validation set} from Rakuten Ichiba dataset.} }
    \vspace{-0.3cm}
    \begin{tabular}{ p{3.8cm}||p{0.9cm}|p{0.9cm}|p{0.9cm}  }
    \hline
    \multicolumn{4}{|c|}{ \textbf{Rakuten Ichiba Product Catalog Dataset} } \\ \hline
    \textbf{Classifiers}& \textbf{Head}  & \textbf{Torso} & \textbf{Tail} \\\hline
    MoE                         & 0.659 & 0.722 & 0.738 \\ \cdashline{1-4}[1pt/1pt]
    Aggregator                  & 0.693 & 0.746 & 0.757 \\ \cdashline{1-4}[1pt/1pt]
    fastText Auto-tuned NNI     & 0.689 & 0.747 & 0.754 \\ \hline
    Ensemble                    & 0.710 & 0.768 & 0.771 \\ \cdashline{1-4}[1pt/1pt]
    MoHE-1                      & 0.715 & 0.770 & 0.771 \\ \cdashline{1-4}[1pt/1pt]
    MoHE-2                      & \textbf{0.722} & 0.770 & 0.773 \\ \hline
    \end{tabular}
    \label{table:Ichiba-macro-F1}
    \vspace{-0.3cm}
\end{table}

Table \ref{table:Ichiba-macro-F1} shows the comparative performance of our proposed model framework against the baselines.
Here too, the model frameworks \texttt{MoHE-2} perform best.
For obtaining the results from SIGIR'18 E-Com dataset, in Table \ref{table:SIGIR18-data-challenge-macro-F1}, the classifiers have been set up as flat classifiers. 
In this case too, \texttt{MoHE-2} outperforms all other models and frameworks compared here.
In Table \ref{tab:baseline_performance_1level}, \texttt{GCP AutoML} shows highest Micro-F1 for this dataset due to a \emph{smaller support set}.

\begin{table}[h!]
    \small
    \centering
    \caption{\small{Macro-F1 scores from the models and frameworks discussed here for the \textit{test set} from SIGIR'18-DC dataset.}}
    \vspace{-0.2cm}
    \begin{tabular}{ p{3.4cm}||p{3.1cm}  }
    \hline
    \multicolumn{2}{|c|}{ \textbf{Rakuten.com SIGIR'18 Data Challenge Dataset} } \\ \hline
    \textbf{Classifiers}& \textbf{Full evaluation data} \\\hline
    MoE                     & 0.392 \\ \cdashline{1-2}[1pt/1pt]
    Aggregator              & 0.443 \\ \cdashline{1-2}[1pt/1pt]
    fastText Auto-tuned NNI & 0.446 \\ \hline 
    Ensemble                & 0.458 \\ \cdashline{1-2}[1pt/1pt] 
    MoHE-1                  & 0.461 \\ \cdashline{1-2}[1pt/1pt] 
    MoHE-2                  & \textbf{0.472} \\ \hline 
    \end{tabular}
    \label{table:SIGIR18-data-challenge-macro-F1}
    \vspace{-0.3cm}
\end{table}

For all results in tables \ref{table:meta-performance}, \ref{table:Ichiba-macro-F1} and \ref{table:SIGIR18-data-challenge-macro-F1}, numbers in bold means that they are significantly better than all other numbers in the same column with statistical significance being measured using bootstrap sampling \cite{Yiu-Chang_et_al-sigir-dc-ICTIR-2019} with $95\%$ confidence interval.

\subsection{Ablation Studies for MoHE Metadata}
\label{subsection:Ablation-studies-for-MoHE}

We now compare our proposed \texttt{MoHE} frameworks with and without the use of metadata features as a conclusion to the discussion in \S\ref{subsection:MoHE-meta-features}.
Based on the ablation studies shown in Table \ref{table:meta-performance}, both ``\texttt{shop\_ID}'' and ``\texttt{tag\_ID}'' turn out to have strong correlations with labels.
Effectiveness of descriptions largely depends on genres, yet including tokens from descriptions with chosen parts of speech improves overall performance. 
By utilizing all three types of metadata, the largest level one genres gain $2-3\%$ macro-F1 performance depending on the framework, and we have observed that some of the tail genres gain more than $10\%$. 
By a design choice, the values for ``\texttt{shop\_ID}'' and ``\texttt{tag\_ID}'' are given to the same metadata thread while description is given to a separate metadata estimator thread.

\begin{table}[h!]
    \small
    \centering
    \caption{\small{Macro-F1 values for the \texttt{MoHE-(1/2)} classifiers without and with metadata for level one genres in the \textit{validation set}. Notations for the added meta data values are meta-1 (\texttt{shop\_ID}), meta-2 (\texttt{shop\_ID}+\texttt{tag\_ID}), and meta-3 (\texttt{shop\_ID}+\texttt{tag\_ID}+\texttt{description}). As noted in \S\ref{subsection:MoHE-meta-features}, method-1 and method-2 are two different ways of adding metadata to \texttt{MoHE} frameworks.}}
    \vspace{-0.3cm}
    \begin{tabular}{ p{3.8cm}||p{0.9cm}|p{0.9cm}|p{0.9cm}  }
    \hline
    \multicolumn{4}{|c|}{ \textbf{Rakuten Ichiba Product Catalog Dataset} } \\ \hline
    \textbf{Classifiers}& \textbf{Head}  & \textbf{Torso} & \textbf{Tail} \\\hline
    %
    MoHE-1, without meta data   & 0.715 & 0.770 & 0.771 \\ \cdashline{1-4}[1pt/1pt]
    MoHE-1, meta-3 (method-1)   & 0.742 & 0.794 & 0.793 \\ \hline
    \hline
    MoHE-2, without meta data   & 0.722 & 0.770 & 0.773 \\ \cdashline{1-4}[1pt/1pt]
    MoHE-2, meta-1 (method-2)   & 0.738 & 0.783 & 0.783 \\ \cdashline{1-4}[1pt/1pt]
    MoHE-2, meta-2 (method-2)   & 0.741 & 0.792 & 0.789 \\ \cdashline{1-4}[1pt/1pt]
    MoHE-2, meta-3 (method-2)   & \textbf{0.745} & \textbf{0.797} & 0.793 \\ \hline
    \end{tabular}
    \label{table:meta-performance}
    \vspace{-0.3cm}
\end{table}


Table \ref{table:meta-performance} shows that for all head, torso and tail segments for $L1$ genres, \texttt{MoHE-2, meta-3 (method 2)} performs best although not statistically significant from \texttt{MoHE-1, meta-3 (method-1)} for the tail segment.

\subsection{Evaluating Models using User Sessions}
\label{subsection:evaluations-user-interactions}

We now direct our attention to the most pressing problem of continuous model evaluation for large scale e-commerce catalog classification without repeatedly having human-in-the-loop annotation.
We use data from user interactions with the catalog, \textit{i.e.} user sessions, to obtain \textit{possible} \textit{End-user Perspective}, or \texttt{EuP} (\textit{pronounced ``Yup''}), scores for a given labeling of a dataset.
In a nutshell, the \texttt{EuP} scores show users' confidence on trusting a labeling of a dataset given the information need reflected in their queries that trigger purchase events.
We say possible since users don't interact will all catalog items.
So, why do we need \texttt{EuP} scores?

\textbf{1.} Users searching for a product \textit{usually} select an item if the genre of the product generally matches that of their queries.
This is especially true for purchased items. 

\begin{figure}[t!]
    \centering
    
	\subcaptionbox{\small{Plot of query length in characters vs. frequency obtained from user sessions. Queries are obtained from sessions corresponding to items \textit{purchased} in the evaluation set.} \label{figure:query-length-vs-frequency-eval-set}}{\includegraphics[width=0.4\columnwidth]{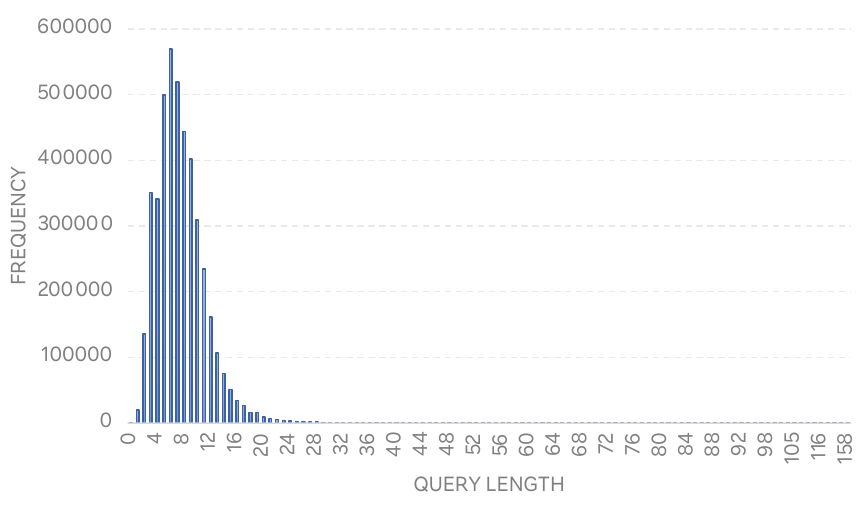}}%
    \hspace{0.05\textwidth}
    \subcaptionbox{\small{X-axis represents time of user interactions with a \textit{particular tail item}. The time span is around one year. The Y-axis is the query character overlap with the item's genre path }\label{figure:query-overlap-after-category-restructuring}}{\includegraphics[width=0.4\columnwidth]{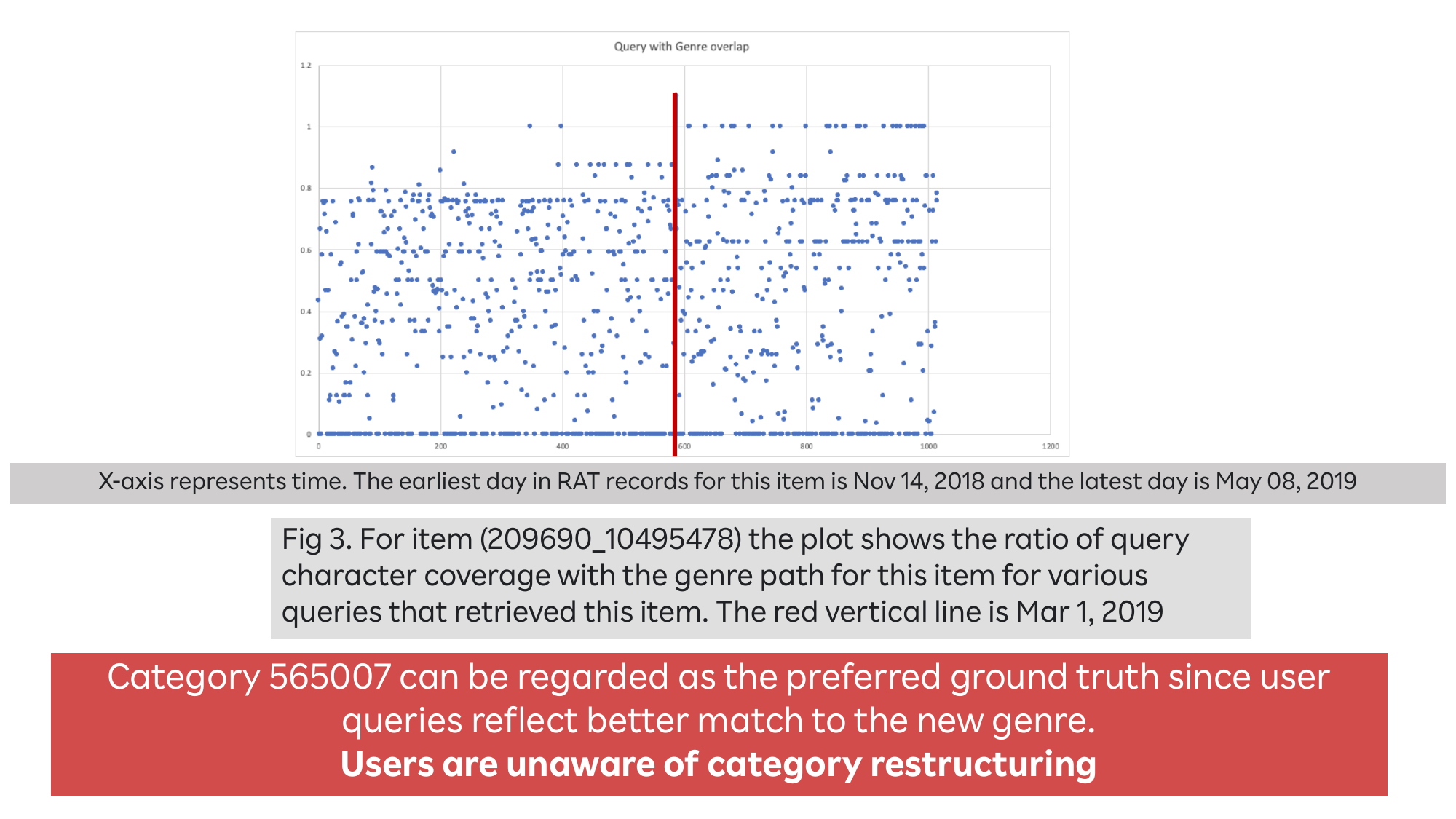}}%
    
    \vspace{-0.2cm}
    \caption{\small{\emph{Left Figure}: $75\%$ of the queries are at least $5$ character long. \emph{Right Figure}: A category restructuring happens at the time where the red vertical line appears. After that time queries with length $> 5$ have more overlap  with the item's genre path for $90\%$ threshold.}} 
    \label{figure:query-overlap-with-genres}
    \vspace{-0.6cm}
\end{figure}

\textbf{2.} Assignment of genres to catalog items are based either on merchant uploads that could be incorrect or genre assignments from previous models trained on data with label noise.
Some labels can change over time based on business needs or fixes by the QA team. 
For instance, Fig. \ref{figure:query-overlap-after-category-restructuring} shows a business need to restructure the category for a tail item after which the item's genre path becomes more relevant to search queries that retrieve the item.
As such, the genre paths for the catalog items may not completely match those that are frozen when model training begins.
The mismatch is much less pronounced for the validation set (that is obtained from the training set) than for an evaluation set that has been curated.

\textbf{3.} Models need to be evaluated using real and updated data, and A/B testing through the primary search interface on a continuous basis is not feasible for \emph{all} or \emph{a majority of} genres.

These observations have led us to review model evaluation using implicit customer feedback and validation.
We aim to find a correlation measure between \texttt{EuP} scores obtained from an \textit{evaluation set} to those from the \textit{validation set} to understand the approximate noise in the labels that will affect model performance.
More specifically, using \texttt{EuP} scores we want to \textbf{quantitatively estimate an approximate amount of label noise} from items that have been purchased for a subset of genres.
We use anonymized user sessions and catalog data made available to us by \texttt{Rakuten Ichiba}.

\paragraph{\textbf{Evaluating Models with Customer Validation}:} The evaluation set is a one-off set of ground truth genres that is sampled from a distribution over GMS and covering all level one genres.
Manual corrections of the genres induce large covariate shift.

For a given dataset (evaluation, validation or even training), we select a subset of items that have been \textbf{purchased} at least once over a period of one year.
For each such session for which there is a query of length $\geq 5$ -- see Fig. \ref{figure:query-length-vs-frequency-eval-set}, we compute overlap of query characters with the genre path names of the items.
Heuristically, the coverage threshold is set to be $> 90\%$ to collect (\textit{provided  genre path} ($\mathbf{y}^+_l$), \textit{item genre path} ($\mathbf{y}_l$)) pairs.
Items in the catalog are labeled with $\mathbf{y}_l$ and are stored as \texttt{genre\_ID}s \textit{for items} (see Fig. \ref{figure:user-session}).
The \textit{provided genre path}s are the ground truth labels for evaluation set and training set labels for the validation set.
Thus for a particular genre path, there are mappings $\mathbf{y}^+_l \rightarrow \{\mathbf{y}_l^{(i)}\}$ with $i$ being an item in the user sessions.
We then compute the accuracy of agreement of $\mathbf{y}^+_l$ to $\{\mathbf{y}_l^{(i)}\}$ and \emph{denote this agreement to be the \texttt{EuP} score for} $\mathbf{y}^+_l$.

Using user session data, out of a total of $\approx 13K$ training genres, the evaluation set has $3{,}731$ and the validation set has $5{,}751$ genres.
Their intersection has $3{,}478$ genre paths on which all \texttt{EuP} scores in Tables \ref{table:auto-evaluation} and \ref{table:auto-validation} are obtained.

\begin{table}[h]
    \small
    \centering
    \caption{\small{Average \texttt{EuP} scores for $3{,}478$ level one genre paths from the out-of-distribution \textbf{evaluation set} that have item purchases. The individual F1 scores are weighted with genre specific \texttt{EuP} scores.}}
    \vspace{-0.3cm}
    \begin{tabular}{ p{3.8cm}||p{0.9cm}|p{0.9cm}|p{0.9cm}  }
    \hline
                            & \textit{Head} & \textit{Torso} & \textit{Tail} \\\hline
    \multicolumn{4}{|c|}{\textbf{Average \texttt{EuP} scores for ground truth labels.}} \\ \hline
    All \texttt{L1} genres  & 0.771 & 0.810 & 0.756 \\ \hline
    \multicolumn{4}{|c|}{\textbf{\texttt{EuP} score weighted F1s}} \\ \hline
    fastText Auto-tuned NNI & 0.465 & 0.500 & 0.477 \\ \hline
    Ensemble                & 0.500 & 0.547 & 0.505 \\ \hline
    MoHE-2                  & 0.492 & 0.541 & 0.501 \\ \hline
    \end{tabular}
    \label{table:auto-evaluation}
    \vspace{-0.3cm}
\end{table}

Table \ref{table:auto-evaluation} shows average \texttt{EuP} scores across all level one genre paths for the head, torso and tail segments.
Since the evaluation set is out-of-distribution, we will expect the scores to be less.
The scores are indeed low and indicative of the $20\%$ label noise as mentioned in \S\ref{section:introduction}, when we compare them to those in Table \ref{table:auto-validation} and we measure them for purchased items only.
For the evaluation set, as expected, \texttt{fastText Autotune NNI} has fitted extremely well to the noisy training set, much more than \texttt{MoHE-2}, whereas \texttt{Ensemble} has handled covariate shift better.
This trend is seen even without the \texttt{EuP} weights, where the scores are $10\%$ higher for all classifiers.
Note that there is no purchased item selection bias while \emph{training models}.

When we consider the validation set that is in-distribution, the \texttt{EuP} scores are much higher since the \textit{provided genre path} labels are from the validation set as shown in Table \ref{table:auto-validation}.
Here, the scores follow the general trend where \texttt{MoHE-2} $>$ \texttt{Ensemble} $>$ \texttt{Fasttext} and are higher here than those shown in Table \ref{table:Ichiba-macro-F1} since the number of genres is much less due to the user interaction and query overlap criterion filter.
From a practical standpoint, reporting meaningful comparison numbers to business heads is critical for project sustenance.
To this end, reporting \texttt{EuP} scores based on categories that customers are interacting with may be better than reporting a full spectrum of ``\textit{science KPI}'' numbers only as shown in Tables \ref{tab:baseline_performance_1level} and \ref{table:Ichiba-macro-F1}.

\vspace{-0.2cm}
\begin{table}[ht]
    \small
    \centering
    \caption{\small{Average \texttt{EuP} scores for $3,478$ level one genre paths from the in-distribution \textbf{validation set} that have item purchases. The individual F1 scores are weighted with genre specific \texttt{EuP} scores.}}
    \vspace{-0.3cm}
    \begin{tabular}{ p{3.8cm}||p{0.9cm}|p{0.9cm}|p{0.9cm}  }
    \hline
                            & \textit{Head} & \textit{Torso} & \textit{Tail} \\\hline
    \multicolumn{4}{|c|}{\textbf{Average \texttt{EuP} scores for validation labels.}} \\ \hline
    All \texttt{L1} genres  & 0.948 & 0.957 & 0.911 \\ \hline
    \multicolumn{4}{|c|}{\textbf{\texttt{EuP} score weighted F1s}} \\ \hline
    fastText Autotuned NNI & 0.741 & 0.759 & 0.727 \\ \hline
    Ensemble                & 0.744 & 0.771 & 0.736 \\ \hline
    MoHE-2                  & 0.750 & 0.772 & 0.738 \\ \hline
    \end{tabular}
    \label{table:auto-validation}
    \vspace{-0.4cm}
\end{table}

The Pearson correlation coefficient between the validation and evaluation \texttt{EuP} scores for the $3,478$ categories is only $0.193$ indicating a significant level of covariate shift in the evaluation data.
To conclude this section, we propose an open problem on defining item reachability for \emph{non-purchased} items in a catalog with the eventual goal of measuring a classifier on expected revenue.
In the interest of space, we refer the interested reader to \S\ref{subsect:evaluations-for-item-reachability}.

%% file: src/conclusion.tex
\section{Conclusion}
\label{section:conclusion}

We propose a lightweight Neural Network ensemble framework for product classification that is adaptable enough to include structured metadata, which are difficult to include in heavyweight language models like \texttt{BERT}.
This novel neural ensemble classification framework preserves the best of ensemble of independent models and model fusion. 
We additionally propose a novel way of measuring label discrepancy between training and evaluation sets using user interactions with the catalog.
To the best of our knowledge, our approach is unique and unlike those found in \cite{Yin_Liangjie-KDD-2019, Wu-et_al-SIGIR-2018, Swaminathan-et_al-NeurIPS-2017}.

%% file: src/supplementary-materials.tex
\section{Supplementary Materials}
\label{section:supplementary-materials}

\begin{figure}[h!]
    \caption{\small{L1 histogram.}}
    \includegraphics[width=\columnwidth]{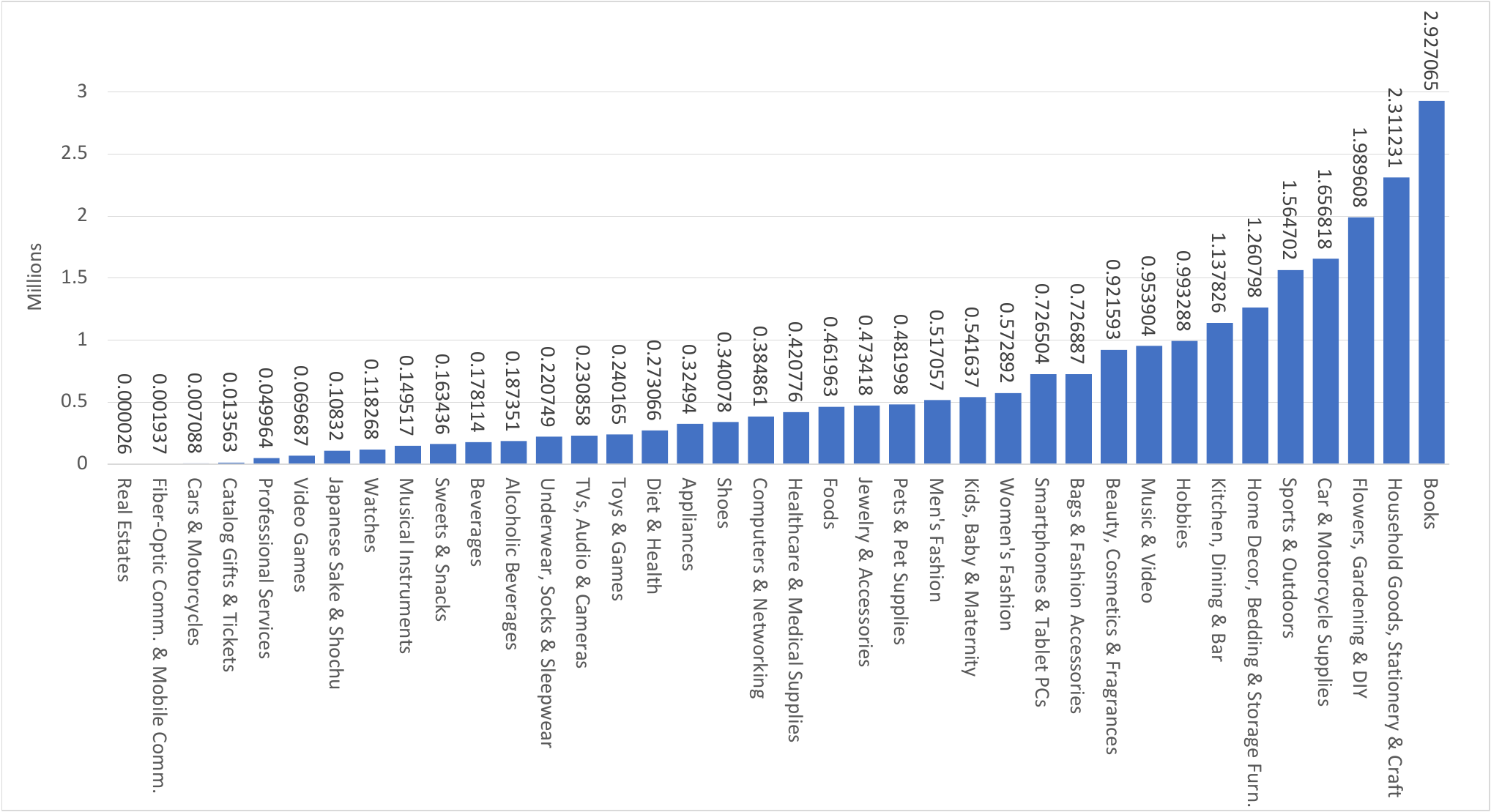}
    \label{figure:L1-histogram-in-millions}
    \vspace{-0.5cm}
\end{figure}

For Ichiba, model is cl-tohoku/bert-base-japanese-whole-word-masking from HuggingFace (Parameter size: 110 M)
For SIGIR, model is bert-base-uncased from huggingface(Parameter size: 110 M)
They are all the base model, which are 12 layers.
 
For more information:
https://huggingface.co/cl-tohoku/bert-base-japanese-whole-word-masking,
https://huggingface.co/bert-base-uncased

\begin{table}[h!]
\centering
\caption{BERT parameter settings.}
\label{table:1}
\begin{tabular}{||c c ||} 
 \hline
 model &  Bert   \\ [0.5ex] 
 \hline\hline
 batch\_size & 128   \\ 
 \hline
 gpus & 4 \\
 \hline
 optimizer & Adam   \\
 \hline
 learning rate & 5e-5   \\
 \hline
 epochs & 40   \\ [1ex]
 \hline
\end{tabular}
\end{table}

\subsection{\textbf{Open Problem}: Evaluating Models for Item Reachability}
\label{subsect:evaluations-for-item-reachability}

For business goals, a final question remains -- what is the impact of \emph{full genre path} classification i.e. do we stop classification at an intermediate level or go down to the leaves? 
\emph{The motivation here is to simulate revenue impact through item classification only}.
``\textit{Item reachability}'' is the degree to which an item in the catalog is reachable from a node in the taxonomy.
We define item reachability from a generative viewpoint but pose its definition as an open problem.

For any internal node $n_l$ of a genre path $\overrightarrow{p_{L}}$, which has $S_l$ siblings at level $l$, a particular item $\mathbf{x}_i \in \{\mathbf{x}_1, ..., \mathbf{x}_N\}$ is reachable with uniform distribution over $[1,N]$ if we ignore all of the $M$ children of node $n_l$. 
Assume that the $M$ children have item counts $[N_1,...,N_M]$.
Instead, if we generate more specific \textit{sub-topics}, i.e. the child genres of $n_l$ and sample the item from the sub-topic, then reachability improves.
The generative story for an item $\mathbf{x}_i$ thus becomes:
\begin{itemize}
    \item Sample a node $n_l$ with some probability of user verification.
    \item Sample a prior over proportions: $ \theta_m \sim Dirichlet(\boldsymbol{\alpha})$
    \item Sample a child $m$ of $n_l \sim Mult\left(\theta_m \right)\,:\, m \in \text{children of}\, n_l$
\end{itemize} 

Algo. \ref{algorithm:item-reachability-evaluation} produces an ``\textit{Item Reachability}'' score for each item $\mathbf{x}$ in a held-out (here validation) set that has \textbf{not} been purchased or ``added-to-cart''.
The score is defined for its genre path ending at node $n_l$.
Instead of generating a child $c_m$ of $n_l$, we use the \textit{provided genre path} till level $L$, $\mathbf{y}^+_{L}$ as a guide to select $c_m$ and hence $\theta_m$ and $\alpha_m$.
For estimating Dirichlet parameters in Algo. \ref{algorithm:dirichlet-parameter-estimation}, we use the open source implementation\footnote{\url{https://github.com/ericsuh/dirichlet}} of Thomas Minka's paper \cite{Minka-2003}.
The $20\%$ label noise used in estimating proportions and shown in line $8$ is distributed randomly over the other child genres.

\begin{algorithm}[h!]
    \small
    \SetKwInOut{Input}{input}
    \SetKwInOut{Output}{output}
    
    \Input{Set of genre paths $\{[n_0>...>n_L]\}$, evaluation level $l$, predictions from classifier $g$, \texttt{EuP} scores for nodes in the genre paths from validation set.}
    \Output{List of item reachability scores for every node $n_l$}
    
    \KwData{$\mathbf{X}$: The training set.}
    \KwData{$\mathbf{X}_{ho}$: The held-out set {\footnotesize{{for items not purchased or added-to-cart}}}}
    
        $\boldsymbol{\theta} = estimate\_theta(\mathbf{X})$ ; $\,\,$ {\footnotesize{// \textit{Use Maximum Likelihood}}} \;
        $\boldsymbol{\alpha} = estimate\_alpha(\mathbf{X})$; $\,\,$ {\footnotesize{// \textit{Use Algo. \ref{algorithm:dirichlet-parameter-estimation}}}} \;
        
        $i\_r \longleftarrow \texttt{dict}$ \;
        \ForEach {sample $\mathbf{x}$ of the held-out set $\mathbf{X}_{ho}$} {
    
            $f_{g}(\mathbf{x}) = \mathbb{1}_{[g(\mathbf{x})_l==\mathbf{y}^+_l]} \times \widehat{\texttt{EuP}}(\mathbf{y}^+_l)$ ; {\footnotesize{// \textit{normalized \texttt{EuP} over level $l$} }} \;
            $f_{c_m}(\mathbf{x}) = \theta_{c_m}$ ; {\footnotesize{// \textit{$c_m$ is the child of $n_l$}}} \;
            $f_{\theta_m}(\mathbf{x}) = \theta_{c_m}^{\alpha_{c_m}-1} $ \;
    
            $i\_r[c_m] \longleftarrow [i\_r[c_m];\, f_{g}(\mathbf{x}) \times f_{c_m}(\mathbf{x}) \times f_{\theta_m}(\mathbf{x})]$
        }
        \Return $i\_r$
    \caption{Item Reachability}
    \label{algorithm:item-reachability-evaluation}
\end{algorithm}

%
%

\begin{algorithm}[ht!]
    \small
    \SetKwInOut{Input}{input}
    \SetKwInOut{Output}{output}
    
    \Input{A selection of the training dataset $\mathbf{X}$, labeled using $\mathbf{y}^+$. The evaluation level $l$ to obtain $\mathbf{y}^+_l$.}
    \Output{Dirichlet parameters $\boldsymbol{\alpha}$ for every genre node $n_l$.}
    
    
        \ForEach {sample $\mathbf{x}_i$ from $\mathbf{X}$} {
            $k^{(n_l)} \longleftarrow $ genre for node $n_l$ from $\mathbf{y}^+_l$ \;
            \uIf{$\mathbf{x}_i$ has been \emph{``added to cart''} or \emph{``purchased''}}{
                $\eta_{i,k^{(n_l)}} \longleftarrow 0.95$; $ \,\,$ $\eta_{i, \neg k}^{(l)} \longleftarrow \boldsymbol{\zeta}$; {\footnotesize{// \textit{random subdivision of} $0.05$}} \;
            }
            \Else{
                $\eta_{i,k^{(n_l)}} \longleftarrow 0.8 $ ; {\footnotesize{// \textit{accounting for $20\%$ label noise}}} \;
                $ran \longleftarrow \texttt{normalize}_{S-1}\left(\texttt{rand}(S-1)\right) \times 0.2$; $j=0$ \;
                \ForEach{$z \neq k$ with $z = 1, ..., S_l$}{
                    $\eta_{i, z^{(n_l)}} \longleftarrow ran[j++]$ \;
                }
            }
        }
        \Return $\boldsymbol{\alpha} \longleftarrow $ estimate\_Dirichlet($\{k^{(n_l)}:[\boldsymbol{\eta}_{k^{(n_l)}}^{\texttt{T}}]\}$) \;
    \caption{Estimating Dirichlet Parameters}
    \label{algorithm:dirichlet-parameter-estimation}
\end{algorithm}

\begin{table}[h!]
    \small
    \centering
    \caption{\small{Item Reachability from levels three and four. $\mathbf{X}_{ho}$ consists of items in the \textit{validation set} from the $3,478$ genre paths that are \emph{not} purchased or ``added-to-cart'' and irrespective of our overlap criterion. Reachability improves when classification is performed till level $4$ and is better for \texttt{MoHE-2}. The jump is however better for \texttt{Ensemble}. We leave further analysis as an open problem.}}
    \vspace{-0.3cm}
    \begin{tabular}{ p{1.2cm}||p{2.3cm}|p{2.3cm}|p{1.1cm}  }
    \hline
    Classifiers     & R3:$n_{l=3}>..>n_{L=5}$ & R4:$n_{l=4}>..>n_{L=5}$ & R4-R3 \\\hline
    Ensemble        & 39.790\% & 41.453\%   & $1.663\%$ \\ \hline
    MoHE-2          & 39.910\% & 41.531\%   & $1.621\%$ \\ \hline
    \end{tabular}
    \label{table:item-reachability}
    \vspace{-0.2cm}
\end{table}

Table \ref{table:item-reachability} shows \textit{mean} item reachability scores for \texttt{Ensemble} and \texttt{MoHE-2} classification frameworks from levels $3$ and $4$ to leaves.
Based on our reachability definition for non-purchased items, there is $\approx 2\%$ gain.
Assuming that $80\%$ of $400M$ items are not bought, and that $2\%$ of this is lost in reachability if we consider classification only till level $3$, we lose opportunity for $6.4M$ items.
If the median sale price of such items is $\$10$, then potential revenue lost is $\$64M$ not considering sale volume.
Higher accuracy may not always mean higher revenue, unless we optimize for revenue in training loss.